\documentclass[11pt]{article}

\usepackage[preprint]{acl}

\usepackage{times}
\usepackage{latexsym}
\usepackage[most]{tcolorbox}

\usepackage{subcaption}
\usepackage{amsmath}
\usepackage{bm}
\usepackage{amsfonts}
\usepackage{wrapfig}
\usepackage{booktabs}
\usepackage{multirow}
\usepackage{adjustbox}
\usepackage[table]{xcolor}

\newtheorem{theorem}{\textbf{Theorem}}
\newtheorem{theorem*}{Theorem}

\newtheorem{proposition}{Proposition}

\newtheorem{proof}{Proof}

\usepackage[T1]{fontenc}

\usepackage[utf8]{inputenc}

\usepackage{microtype}

\usepackage{inconsolata}

\usepackage{graphicx}

\title{Unlocking the Pre-Trained Model as a Dual-Alignment\\Calibrator for Post-Trained LLMs}

\author{Anonymous Author(s) \\
Anonymous Institution \\
Anonymous Address \\
\texttt{anonymous@email.com}
}

\author{%
  Beier Luo$^{1}$, ~Cheng Wang$^{2}$, ~Hongxin Wei$^{1}$, ~Sharon Li$^{3}$, ~Xuefeng Du$^{4}$\thanks{Corresponding author (xuefeng.du@ntu.edu.sg)} \\
  $^1$Department of Statistics and Data Science,\\Southern University of Science and Technology\\
  $^2$School of Computing, National University of Singapore\\
  $^3$Department of Computer Sciences, University of Wisconsin-Madison\\
  $^4$College of Computing and Data Science, Nanyang Technological University
}

\begin{document}
\maketitle

\newcommand{\methodname}{\texttt{Dual-Align}}

\begin{abstract}
Post-training improves large language models (LLMs) but often worsens confidence calibration, leading to systematic overconfidence. Recent unsupervised post-hoc methods for post-trained LMs (PoLMs) mitigate this by aligning PoLM confidence to that of well-calibrated pre-trained counterparts. However, framing calibration as static output-distribution matching overlooks the inference-time dynamics introduced by post-training. In particular, we show that calibration errors arise from two regimes: (i) \textit{confidence drift}, where final confidence inflates despite largely consistent intermediate decision processes, and (ii) \textit{process drift}, where intermediate inference pathways diverge. Guided by this diagnosis, we propose \methodname{}, an unsupervised post-hoc framework for dual alignment in confidence calibration. \methodname{} performs \textit{confidence alignment} to correct confidence drift via final-distribution matching, and introduces \textit{process alignment} to address process drift by locating the layer where trajectories diverge and realigning the stability of subsequent inference. This dual strategy learns a single temperature parameter that corrects both drift types without sacrificing post-training performance gains. Experiments show consistent improvements over baselines, reducing calibration errors and approaching a supervised oracle.
\end{abstract}

\begin{figure*}[t]
  \centering
  \begin{subfigure}[t]{1\textwidth}
    \centering
    \includegraphics[width=\linewidth, page=1]{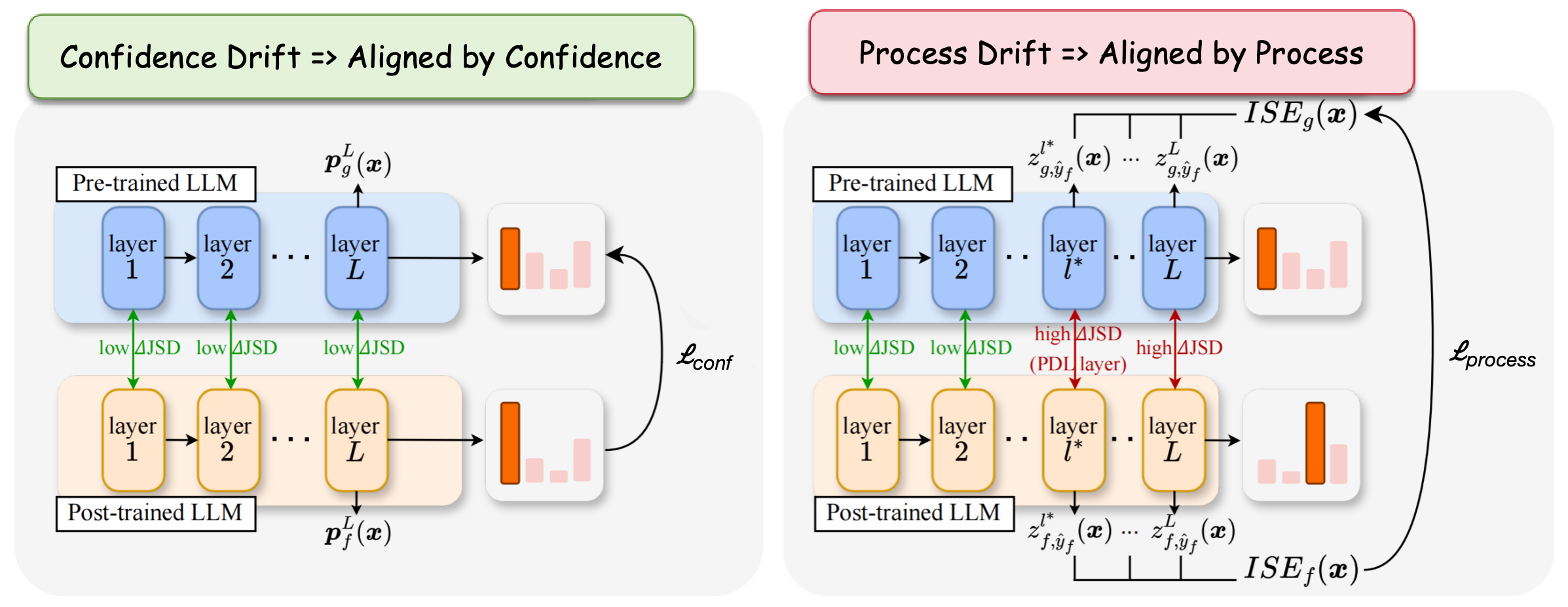}
  \end{subfigure}
    \vspace{-1.5em}
  \caption{\small \textbf{Illustration of our method: DUAL-ALIGN.} Our approach addresses both confidence drift and process drift. For confidence drift, we align the LLMs’ confidence using the objective $\mathcal{L}_{\rm Conf}$ (Left). For process drift, we first identify the Peak Divergence Layer (PDL), then calculate the Inferential Stability Entropy (ISE) with respect to the process drift between the PLM and PoLM, and align it using the objective $\mathcal{L}_{\rm Process}$ (Right).}
    \label{fig:framework}
\end{figure*}

\section{Introduction}

Post-training methods such as instruction tuning and reinforcement learning from human feedback, substantially improves large language model (LLM) alignment and adaptability across tasks \citep{wei2022finetuned,ouyang2022training,zhang2025instructiontuninglargelanguage}. 
Yet it also introduces new challenges in uncertainty estimates, often amplifying over-confidence relative to the pre-trained language models (PLMs) \citep{achiam2023gpt,shen2024thermometer}. 
To circumvent this, researchers have explored post-hoc confidence calibration, such as temperature scaling (TS)~\citep{guo2017calibration} for post-trained LMs (PoLMs): aligning predicted probabilities with empirical accuracy so models behave cautiously under uncertainty \citep{xiong2023can}. 

Recent unsupervised methods, such as DACA~\citep{luo2025your}, use the confidence of the well-calibrated PLM on unlabled data as a reference to calibrate the PoLM. 
To avoid potential conflicts from new knowledge introduced by post-training, DACA chooses to only align on samples where predictions are consistent between PLM and PoLM. 
However, this selective alignment strategy is inherently data-inefficient, as it discards all samples where the models disagree. 
More critically, by focusing solely on matching the final output confidence, it treats calibration as a static, surface-level matching problem. 
This fails to address the complex drifts in the model's intermediate inference process induced by post-training, which are often the root cause of miscalibration. 
We raise a key question here: \textit{How does post-training alter the decision process of LLMs, and can we use that understanding to calibrate them more effectively?}

To answer this, we begin by investigating the different behavioral regimes of the PLM and PoLM by analyzing their differences w.r.t. the layer-wise predictions and final outputs. 
Our analysis at Figure~\ref{fig:jsd_trajectory}  reveals two distinct post-training phenomena: (i) In samples where the PoLM and PLM agree on the final answer, their intermediate decision processes are largely consistent, yet the PoLM's final confidence is systematically inflated—a phenomenon we term \textbf{confidence drift}. (ii) Conversely, in samples where they disagree, the models' decision pathways diverge sharply at a specific intermediate layer, causing their inference trajectories to split and lead to different answers. We term this more fundamental change \textbf{process drift}. These observations motivate a calibration approach that addresses both phenomena at their source.

\paragraph{Our contributions.} 
To this end, we propose \methodname{}, a novel post-hoc LLM calibration framework (Figure \ref{fig:framework}) that treats calibration as a \emph{dual alignment} problem. 
It performs (1) \textbf{confidence alignment} to correct surface-level overconfidence by matching the PoLM’s final-layer output distribution with the PLM's. Our motivation for pursuing deeper alignment in the models' inference process arises from a key observation: post-training creates a problematic pattern in which extreme overconfidence is coupled with unnaturally low Inferential Stability Entropy (ISE) (Figure~\ref{fig:ise}), calculated over the LLM inference trajectory through different layers. 
To rectify this, we introduce a novel (2) \textbf{process alignment}, which first identifies the Peak Divergence Layer (PDL)—the point at which the inference pathways of the PLM and PoLM models diverge most significantly—and then aligns the PoLM’s ISE with the PLM’s healthier distribution from that layer forward. 
Importantly, our framework interpolates between these two objectives on a per-sample basis using a divergence-derived weight coefficient. 
This approach produces a temperature parameter that adapts to different miscalibration regimes, while \textit{preserving the performance gains achieved through post-training}.
Both theoretical results (Proposition~\ref{prop:process_align_calibration}) and empirical findings (Section~\ref{sec:experiments}) demonstrate that \methodname{} achieves substantial improvements, reducing the Expected Calibration Error (ECE) by more than 30\% across various advanced LLM architectures compared to strong baselines.

\section{Preliminaries}

\paragraph{Probability distribution across transformer layers.}
Formally, we define the input prompt as a sequence of tokens $\bm{x} = \{x_1, x_2, \dots, x_N\}$ and our analysis focuses on the final token, $x_N$, as its hidden state is used to generate the model's prediction. At each layer $l \in [1, L]$ of a transformer model \citep{vaswani2017attention}, the hidden state for this token is conceptually updated as:
\begin{equation}
\small
\bm{h}^l(x_N) = \bm{h}^{l-1}(x_N) + \text{Attn}^l(x_N) + \text{MLP}^l(x_N),
\end{equation}
where $\bm{h}^l \in \mathbb{R}^{d_{\text{model}}}$ denotes the hidden state at the $l$-th layer. Using LogitLens \citep{nostalgebraist2020logitlens}, we can project any intermediate hidden state $\bm{h}^{l}(x_N)$ into the vocabulary space via the unembedding matrix $W_U \in \mathbb{R}^{V \times d_{\text{model}}}$, with $V$ as the vocabulary size. Since the embedding $\bm{h}^{l}(x_N)$ encapsulates information from the entire input $\bm{x}$, we denote the resulting per-layer logits as 
\begin{equation}\bm{z}^{l}(\bm{x}) = W_U \cdot\bm{h}^{l}(x_N) \in \mathbb{R}^{V}.\end{equation}

Our analysis primarily focuses on Multiple-Choice Question Answering (MCQA) problems, which typically present a set of options, such as $\mathcal{Y} = \{A, B, C, D\}$. The probability of each option at layer $l$ is given by
\begin{equation}
p_i^l(\bm{x}) = \frac{\exp(z^l_i(\bm{x}))}{\sum_{j \in \mathcal{Y}} \exp(z^l_j(\bm{x}))}, \quad i\in\mathcal Y.
\end{equation}

\paragraph{Confidence calibration for PoLMs.}
We aim to calibrate a post-trained language model PoLM, denoted by $f$, using a pre-trained language model PLM, $g$, as a reference. In the context of a multiple-choice question, the model's prediction, $\hat{y}_f(\bm{x})$, is the choice with the highest probability at the final-layer $L$, and this maximum probability value is taken as its confidence, $\hat{P}(\bm{x})=\underset{i\in\mathcal Y}{\max}\,p_i^L(\bm{x})$. A model is considered perfectly calibrated if its confidence matches its true accuracy, i.e., $\Pr\big(Y=\hat{y}\,\big|\,\hat{P}=\beta\big) = \beta$, where $Y$ is the ground-truth. 

A standard metric to measure this discrepancy is the Expected Calibration Error (ECE)~\citep{naeini2015obtaining}. In practice, ECE is estimated empirically by partitioning $K$ samples into $M$ bins ${b_1, b_2, \dots, b_M}$ based on the model's predicted confidence scores, and then computed as:
\begin{equation}
\mathrm{ECE} = \sum_{m=1}^M \frac{|b_m|}{K}\,\big|\mathrm{acc}(b_m)-\mathrm{conf}(b_m)\big|,
\end{equation}
where $\mathrm{acc}(b_m)$ and $\mathrm{conf}(b_m)$ are the average accuracy and confidence in bin $b_m$. A smaller ECE indicates better calibration performance of the model. While PLMs are often well-calibrated, literature recognize that post-training often degrades this property, leading to overconfident predictions~\citep{xiao2025restoring, luo2025your, leng2025tamingoverconfidencellmsreward}, as shown in Figure~\ref{fig:reliability}.

\paragraph{Post-hoc calibration methods.}
Post-hoc calibration adjusts a model's confidence without altering its predictions. A popular supervised method is Temperature Scaling (TS)~\citep{guo2017calibration}, which softens the probability distribution by applying a scalar temperature $\tau > 0$ to the final-layer logits:
\begin{equation}
    p(y=j \mid \bm{x}, \tau) = \mathrm{softmax}\!\left(\frac{\bm{z}_{j}^{L}(\bm{x})}{\tau}\right).
\end{equation}
The temperature $\tau$ is optimized on a labeled dataset. Since the parameter $T$ does not alter the maximum value of the softmax function, the predicted class remains the same. In other words, \textit{temperature scaling does not affect the model’s accuracy}. 

To eliminate the need for labels in calibration, unsupervised methods like DACA~\citep{luo2025your} align the PoLM's confidence with that of the better-calibrated PLM. Crucially, DACA performs this alignment exclusively on samples where the models agree on the prediction, thereby avoiding under-confidence issues caused by optimizing on disagreement cases. However, it treats calibration as a static, surface-level matching problem. This fails to address the complex drifts in the model's intermediate inference process induced by post-training, which motivates our paper.

\section{Understanding the Effects of Post-training}

To investigate how post-training influences a model’s calibration behavior during inference across different layers, we first quantify the changes in predictive distributions at each layer following post-training. Specifically, we measure the divergence $d^l(\bm{x})$ between the pre-trained and post-trained models using the Jensen-Shannon Divergence (JSD), defined as $d^l(\bm{x}) = D_{JS}(\bm{p}_g^{l}(\bm{x})~||~\bm{p}_f^{l}(\bm{x}))$. 

\begin{figure}
    \centering
    \includegraphics[width=1\linewidth]{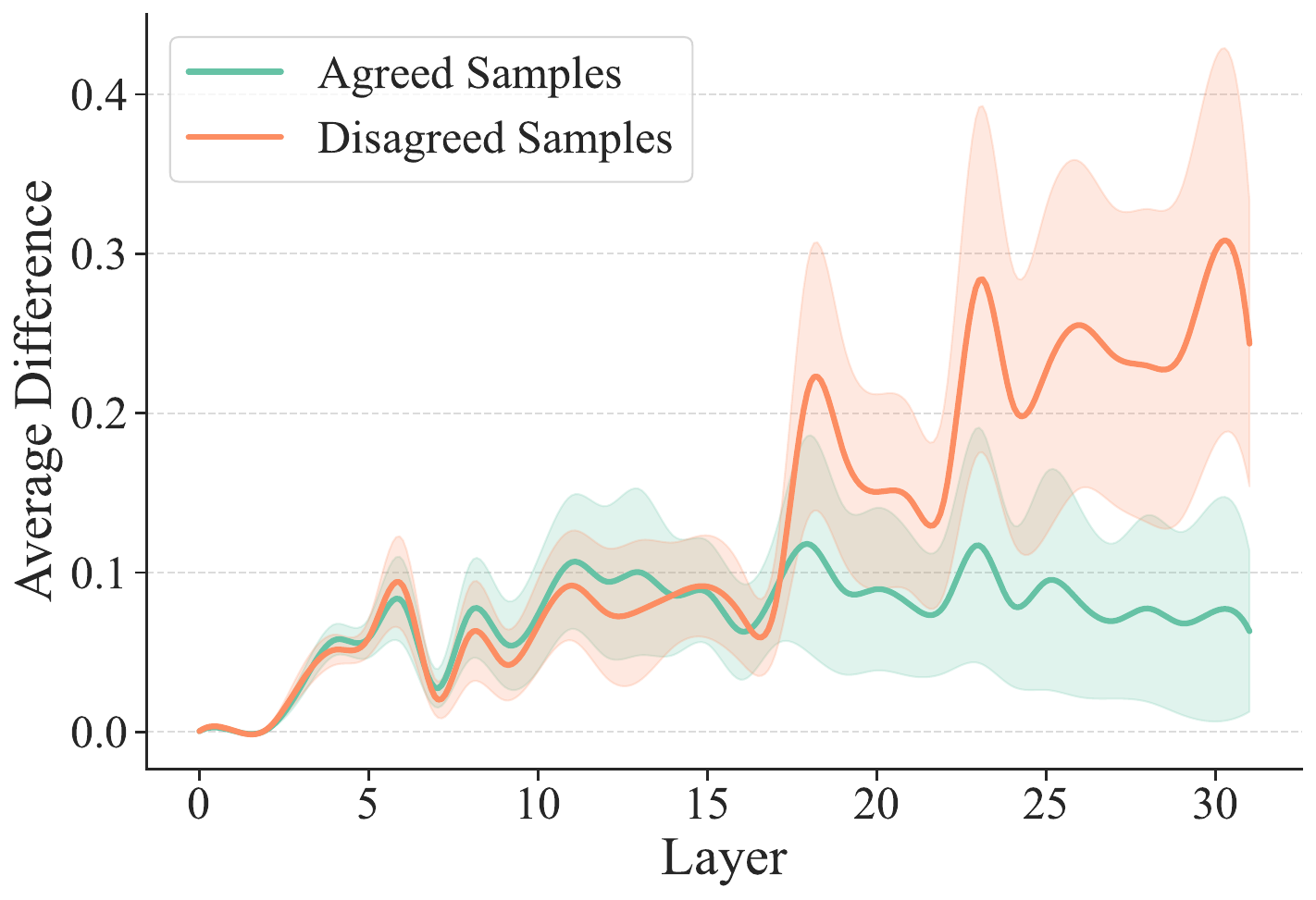}
    \vspace{-1.2em}
    \caption{\small \textbf{The layer-wise Jensen-Shannon Divergence between a post-trained mode Llama-3.1-8B-Instruct and a pre-trained model Llama-3.1-8B on MMLU.} Agreed samples show minimal differences, suggesting \textit{confidence drift}, while disagreed samples display a sharp spike at an intermediate layer, indicating \textit{process drift}.}
    \vspace{-1em}
    \label{fig:jsd_trajectory}
\end{figure}

Surprisingly, we observe that, regardless of whether the PLM and PoLM ultimately produce different predictions, the divergence \( d^l(\mathbf{x}) \) between their predictive distributions remains negligible in the early layers. As illustrated in Figure~\ref{fig:jsd_trajectory}, for samples on which the two models disagree, the divergence \( d^l(\mathbf{x}) \) exhibits a sharp increase at a specific intermediate layer. We refer to this layer as the \textbf{Peak Divergence Layer (PDL)}, defined as
\begin{equation}
l^*(\mathbf{x}) = \underset{l \in \{2, \dots, L\}}{\arg\max} \Big( d^l(\mathbf{x}) - d^{l-1}(\mathbf{x}) \Big).
\end{equation}
Intuitively, PDL corresponds to the earliest layer where post-training induces a qualitative change in the inference dynamics, analogous to a bifurcation point in dynamical systems \citep{kuznetsov1998elements}.

\paragraph{Confidence drift.}
We define \textit{Confidence Drift} as the overconfidence observed in agreement samples, where the intermediate decision process of the PoLM remains consistent with that of the PLM, but the output confidence level is inflated. This phenomenon occurs without any significant change in the decision-making process itself, leading to an exaggeration of the model's certainty.

\paragraph{Process drift.}
In contrast to confidence drift, \textit{Process Drift} refers to the divergence between the prediction distributions of the PLM and PoLM following the PDL $l^*$ on disagreement samples. Specifically, process drift occurs when there is a notable deviation in the intermediate decision processes between the PLM and PoLM, resulting in a different final prediction. Previous research \citep{luo2025your} has shown that confidence alignment on agreement samples can mitigate confidence drift; however, it does not address the root cause of process drift, which remains a crucial aspect to understand in post-training adjustments.

\begin{figure*}[t]
  \centering
  \begin{subfigure}[t]{0.47\textwidth}
    \centering
    \includegraphics[width=\linewidth, page=1]{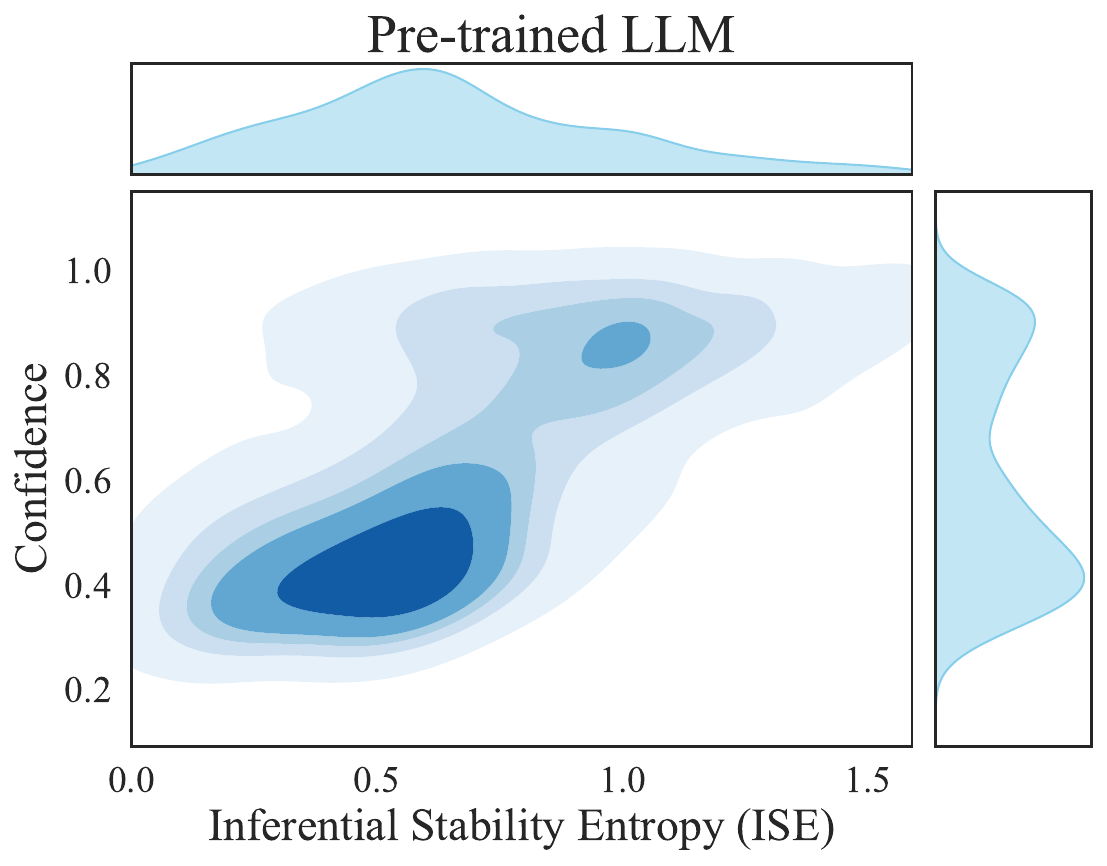}
  \end{subfigure}
  \hfill
  \begin{subfigure}[t]{0.47\textwidth}
    \centering
    \includegraphics[width=\linewidth, page=1]{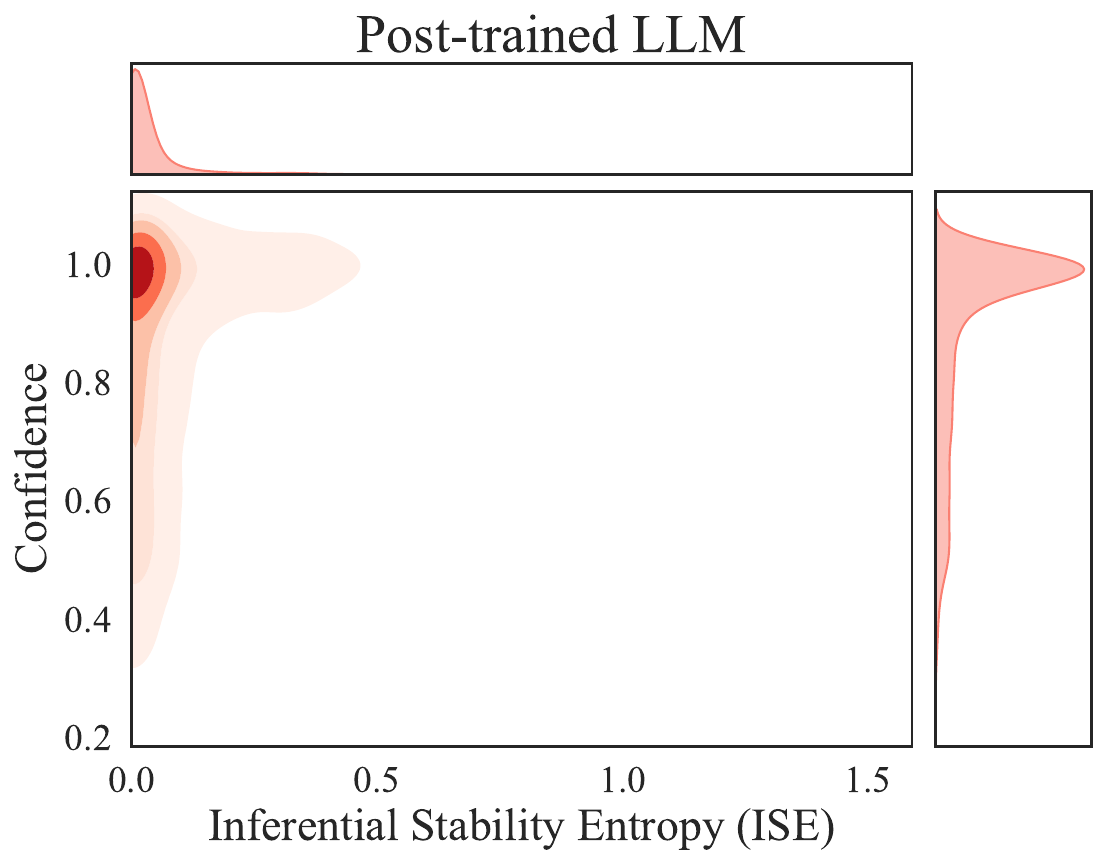}
  \end{subfigure}
    \vspace{-0.8em}
  
  \caption{\small \textbf{Relationship between output confidence and Inferential Stability Entropy (ISE) of Qwen2.5-14B nad Qwen2.5-14B-Instruct on MMLU.} The well-calibrated pre-trained model (left) displays an ISE distribution similar to a normal distribution, whereas the post-trained model (right) shows extreme overconfidence and abnormally low ISE values, indicating overly rigid decision-making processes.}
  \label{fig:ise}
  \vspace{-1.5em}
\end{figure*}

\section{Methodology}

\subsection{Stability of Inference Across Transformer Layers}
A process drift represents a more significant alteration, where the PoLM's intermediate decision process diverges sharply from the PLM's, resulting in a different final answer. 
For such cases, naively aligning the confidence between the PoLM and PLM is counterproductive: it would force the PoLM to match a conclusion produced by a fundamentally different inference process, often resulting in underconfidence \citep{luo2025your}.
Instead, our key insight is to regularize the PoLM’s intermediate inference process itself. Specifically, we propose aligning the \textit{stability} of the model’s inference after the point of divergence. This approach ensures that even when the PoLM arrives at a different conclusion, its confidence in that conclusion mirrors the well-calibrated and stable certainty of a PLM, thereby preventing erratic overconfidence.

To measure the conviction stability of LLMs, we define the \textbf{Inferential Stability Entropy (ISE)} after the PDL $l^*$ as
\begin{equation}
\text{ISE}(\bm{x}) = - \sum_{l=l^*}^{L} q^l(\bm{x}) \log q^l(\bm{x}),  
\end{equation}
where \begin{equation}
\small
q^{l}(\bm{x}) = \frac{\exp(v^{l}(\bm{x}))}{\sum_{j=l^*}^{L} \exp(v^{j}(\bm{x}))}\quad l \in \{l^*,\ldots,L\},\end{equation}
and $v^{j}(\bm{x})$ is the logit of the predicted token at layer $j$. Intuitively, the ISE measures how concentrated the model’s inferential conviction is across layers after the divergence point. The normalized weights $q^{l}(\bm{x})$ form a distribution over layers, indicating where the decision is most strongly formed. A lower ISE corresponds to a sharply peaked distribution, meaning the model commits to a conclusion early and maintains a rigid, homogeneous confidence thereafter, whereas a higher ISE reflects a more gradual and stable consolidation of inference across layers.

\subsection{Process Alignment for Process Drift}
The key idea of our method is grounded in the hypothesis that the overconfidence exhibited by a PoLM arises from an overly rigid conviction process. Specifically, unlike the more deliberative PLM, a PoLM tends to settle on a decision prematurely and maintain uniformly high confidence throughout its intermediate layers. Under this view, a lower ISE indicates a more homogeneous and inflexible conviction trajectory across layers. This hypothesis is empirically supported by the observations presented in Figure~\ref{fig:ise}.

Specifically, we first note that the well-calibrated PLM’s output confidence spans a reasonably wide range, reflecting a healthy degree of epistemic uncertainty (\textcolor[RGB]{15,92,165}{Left}). In stark contrast, the PoLM exhibits severe overconfidence, with confidence scores overwhelmingly concentrated near 1.0 (\textcolor{red}{Right}). Moreover, the two models display fundamentally different relationships between confidence and inferential stability. For the PLM, confidence remains largely invariant across its typical ISE range, suggesting a decoupling between confidence magnitude and layer-wise stability. Conversely, the PoLM shows an undesirable correlation in which extreme confidence is systematically associated with abnormally low ISE values. This pattern indicates that the PoLM’s conviction process has become excessively certain and exhibits minimal variation across layers. Such behavior is visually reflected in Figure~\ref{fig:ise} by the dense clustering of data points in the top-left region of the plot, where confidence approaches 1.0 as ISE converges toward zero.

This sharp contrast between PoLM and PLM reveals that simply correcting the final output confidence may be insufficient. A better approach is to address the intermediate inference dynamics, which makes the PLM's healthier ISE distribution an ideal target. Our process alignment loss is therefore designed to restore a more stable conviction process for PoLM by minimizing the squared difference between the ISE of the two models:
\begin{equation}
\mathcal{L}_{\rm Process}(\tau; \bm{x}) = \big( \text{ISE}_f(\bm{x}, \tau) - \text{ISE}_g(\bm{x}) \big)^2,
\end{equation}
where we divide the PoLM logits by a temperature $\tau$ to calculate  $\text{ISE}_f(\bm{x}, \tau)$. This objective optimizes $\tau$ to align the stability of the PoLM's inference process with that of a better-calibrated PLM.

\subsection{\methodname{}: A Unified Calibration Framework}

Based on the preceding analysis, we propose \methodname{}, a unified framework that addresses both confidence drift and process drift via confidence alignment and process alignment, respectively. Specifically, we quantify the severity of process drift at the sample level using the magnitude of the peak increase in Jensen–Shannon divergence (JSD), $\Delta {D}_{JS}^{l^*}(\bm{x}) = {D}_{JS}(p_f^{l^*}(\bm{x})~||~p_g^{l^*}(\bm{x})) - {D}_{JS}(p_f^{l^*-1}(\bm{x})~||~p_g^{l^*-1}(\bm{x}))$, which serves as a natural indicator of how sharply the inference processes of the PoLM and PLM diverge for a given input. Similar as DACA \citep{luo2025your}, we adopt the KL divergence for confidence alignment and the loss can be written as
\begin{equation}
    \mathcal{L}_{\rm Conf}(\tau; \bm{x}) = D_{KL}(\bm{p}_g^{L}(\bm{x})~||~\bm{p}_f^{L}(\bm{x}, \tau)).
\end{equation}
Therefore, the final learning objective is a weighted combination of the confidence and process alignment components:
\begin{equation}\label{eq:dual_loss}
\small
\begin{split}
\mathcal{L}_{\text{Dual}}(\tau; \bm{x})
&=
\bigl(1 - \Delta D_{\mathrm{JS}}^{l^*}(\bm{x})\bigr)\,
\mathcal{L}_{\mathrm{Conf}}(\tau; \bm{x}) \\
&\quad+
\Delta D_{\mathrm{JS}}^{l^*}(\bm{x})\,
\mathcal{L}_{\mathrm{Process}}(\tau; \bm{x}) .
\end{split}
\end{equation}

\begin{table*}[t]
\centering
\small
\setlength{\tabcolsep}{2pt}  
\renewcommand{\arraystretch}{0.8}

\begin{adjustbox}{max width=\textwidth}
\begin{tabular*}{\textwidth}{@{\extracolsep{\fill}}clcccc}
\toprule
\multirow{2}{*}{\textbf{Models}} 
& \multirow{2}{*}{\textbf{Methods}} 
& \multicolumn{4}{c}{\textbf{Evaluation Metrics} $\downarrow$} \\
\cmidrule(lr){3-6}
& & \textbf{ECE (\%)} & \textbf{MCE (\%)} & \textbf{ACE (\%)} & \textbf{Brier} \\
\midrule

\multirow{8}{*}{\rotatebox{90}{\textbf{Llama3.1-8B}}}
& Vanilla      & $10.806 \pm 0.275$ & $18.602 \pm 0.212$ & $11.809 \pm 0.652$ & $0.461 \pm 0.005$ \\
& CAPE         & $12.567 \pm 0.134$ & $20.788 \pm 0.841$ & $13.134 \pm 0.257$ & $0.495 \pm 0.001$ \\
& Elicitation  & $13.203 \pm 0.067$ & $40.983 \pm 4.065$ & $21.300 \pm 1.714$ & -- \\
& IC           & $11.716 \pm 0.248$ & $64.448 \pm 29.949$& $19.517 \pm 3.165$ & -- \\
& DACA         & $7.811 \pm 0.619$  & $13.824 \pm 0.667$ & $8.064 \pm 0.544$  & $0.451 \pm 0.004$ \\
& \textbf{\methodname~(Ours)}
               & \bm{$2.871 \pm 0.308$}
               & \bm{$5.587 \pm 0.648$}
               & \bm{$3.222 \pm 0.306$}
               & \bm{$0.445 \pm 0.004$} \\
\cmidrule(lr){2-6}
& TS$^\dagger$ (oracle)
               & $1.526 \pm 0.450$
               & $4.790 \pm 1.090$
               & $1.985 \pm 0.609$
               & $0.441 \pm 0.004$ \\
\midrule

\multirow{8}{*}{\rotatebox{90}{\textbf{Qwen2.5-14B}}}
& Vanilla      & $16.735 \pm 0.375$ & $32.406 \pm 0.583$ & $21.848 \pm 1.130$ & $0.388 \pm 0.006$ \\
& CAPE         & $18.022 \pm 0.061$ & $36.091 \pm 0.501$ & $20.987 \pm 0.340$ & $0.407 \pm 0.001$ \\
& Elicitation  & $15.321 \pm 0.002$ & $85.556 \pm 0.000$ & $31.973 \pm 2.713$ & -- \\
& IC           & $32.852 \pm 0.258$ & $47.360 \pm 5.427$ & $22.089 \pm 0.627$ & -- \\
& DACA         & $5.146 \pm 0.340$  & \bm{$8.867 \pm 0.590$} & $4.427 \pm 0.287$ & $0.329 \pm 0.004$ \\
& \textbf{\methodname~(Ours)}
               & \bm{$2.423 \pm 0.070$}
               & $11.241 \pm 2.918$
               & \bm{$3.602 \pm 0.642$}
               & \bm{$0.326 \pm 0.005$} \\
\cmidrule(lr){2-6}
& TS$^\dagger$ (oracle)
               & $2.297 \pm 0.124$
               & $11.411 \pm 2.996$
               & $3.986 \pm 0.994$
               & $0.326 \pm 0.005$ \\
\midrule

\multirow{8}{*}{\rotatebox{90}{\textbf{Gemma-3-27B}}}
& Vanilla      & $23.842 \pm 0.336$ & $58.230 \pm 8.103$ & $35.240 \pm 2.461$ & $0.481 \pm 0.007$ \\
& CAPE         & $19.891 \pm 0.053$ & $38.791 \pm 0.334$ & $23.281 \pm 0.345$ & $0.445 \pm 0.010$ \\
& Elicitation  & $18.413 \pm 0.284$ & $26.526 \pm 2.564$ & $22.456 \pm 1.326$ & -- \\
& IC           & $36.667 \pm 0.313$ & $53.937 \pm 0.414$ & $36.746 \pm 0.346$ & -- \\
& DACA         & $16.842 \pm 0.324$ & $35.205 \pm 0.660$ & $23.985 \pm 0.524$ & $0.406 \pm 0.006$ \\
& \textbf{\methodname~(Ours)}
               & \bm{$5.247 \pm 0.310$}
               & \bm{$18.065 \pm 8.913$}
               & \bm{$9.175 \pm 1.565$}
               & \bm{$0.379 \pm 0.005$} \\
\cmidrule(lr){2-6}

& TS$^\dagger$ (oracle)
               & $5.225 \pm 0.254$
               & $18.069 \pm 9.148$
               & $8.871 \pm 1.561$
               & $0.359 \pm 0.005$ \\
\bottomrule
\end{tabular*}
\end{adjustbox}
\vspace{-0.5em}

\caption{\small
\textbf{Main evaluation results on MMLU across different LLMs.}
Lower values indicate better performance.
Best results among unsupervised methods are highlighted in \textbf{bold}.
IC denotes \emph{Internal Consistency}, TS denotes \emph{Temperature Scaling}.
$\dagger$ indicates methods with access to labeled data.
Results are averaged over three runs.
}
\label{tab:main_results}
\vspace{-2em}
\end{table*}

This unified objective \footnote{We adopt base-2 logs in JSD calculation to ensure its $\Delta {D}_{JS} \leq$ 1. } uses the model's intermediate predictive divergence $\Delta {D}_{JS}^{l^*}(\bm{x})$ as a data-driven weight  during training. In this way, the loss  dynamically balances the two alignment objectives for each sample, without introducing separate hyperparameter. By minimizing the expected loss $\mathbb{E}_{\bm{x} \in \mathcal{D}}[\mathcal{L}_{\methodname{}}(\tau; \bm{x})]$ over an unlabeled dataset $\mathcal{D} = \{\bm{x}_i\}_{i=1}^{K}$, \methodname{} learns an optimal temperature $\tau^*$ that can comprehensively handle the post-training effects on LLM calibration. 

\noindent\textbf{Remark.} During inference, we apply the learned $\tau^*$ to calibrate PoLMs in their final outputs, which does not require additional computational cost or access to PLMs. Additionally, our method \methodname{} is \textit{post-hoc} and does \textit{not} change the maximum of the softmax function and therefore the token prediction. Model accuracy and the capability introduced by post-training are thus not affected.

\noindent\textbf{Mathematical analysis.}
We provide an intuitive interpretation on why process alignment improves calibration.
For easier analysis, we follow~\cite{guo2025sample} and study a smooth calibration surrogate by approximating predictive confidence with the {logit gap}.
Let $\hat y=\arg\max_{i\in\mathcal Y} p_i^{L}(\bm{x})$ be the PoLM prediction and define
$\Delta(\bm{x})=\bm{z}^{L}_{\hat y}(\bm{x})-\log\!\sum_{j\in\mathcal{Y}\setminus\{\hat y\}}\exp(\bm{z}^{L}_{j}(\bm{x}))$.
Temperature scaling with $\tau>0$ yields a confidence proxy $c_{\tau}(\bm{x})=\sigma(\Delta(\bm{x})/\tau)$, where $\sigma(\cdot)$ indicates the sigmoid function.
We measure calibration via the squared surrogate
$\mathcal{E}_f(\tau):=\mathbb{E}_{\bm{x}}\big[(c_\tau(\bm{x})-\Pr(Y{=}y\mid \bm{x}))^2\big]$ where $y$ is true answer.
Concretely, we have the following proposition.

\begin{proposition}
\label{prop:process_align_calibration}
(Informal). Under mild regularity conditions (Appendix~\ref{app:theory}), there exist bounded weights $w(\bm{x})\ge 0$ such that
\begin{equation}
    \small
\mathcal{E}_f(\tau)
\;\le\;
\mathbb{E}_{\bm{x}}\!\left[w(\bm{x})\big(\mathrm{ISE}_f(\bm{x},\tau)-\mathrm{ISE}_g(\bm{x})\big)^2\right]
\;+\;C_{g},
\end{equation}
where $\mathrm{ISE}_f(\bm{x},\tau)$ is the PoLM inferential stability entropy under temperature $\tau$, $\mathrm{ISE}_g(\bm{x})$ is the PLM stability reference, and $C_g$ is a positive constant relevant to the PLM.
\end{proposition}

\noindent\textbf{Interpretation.}
Proposition~\ref{prop:process_align_calibration} shows that, up to a PLM-dependent constant $C_g$, the PoLM’s calibration surrogate $\mathcal{E}_f(\tau)$ is upper-bounded by the ISE mismatch.
This explains why our process alignment improves calibration. All assumptions and proofs are deferred to Appendix~\ref{app:theory}.

\section{Experiments}

\label{sec:experiments}
\subsection{Experimental Setup}
\label{sec:exp-setup}
\noindent\textbf{Models, datasets and evaluation.}
Our evaluation comprehensively assesses a diverse array of large language models, encompassing various scales and architectures, including the Llama-3.1 series \citep{grattafiori2024llama}, the Gemma-3 series \citep{team2025gemma} and the Qwen-2.5 series \citep{yang2024qwen2}. More details about these LLMs are presented in Appendix~\ref{app:model_details}.

We validate our methodology's efficacy across three widely-adopted evaluation benchmarks: MMLU \citep{mmlu}, and MedMCQA \citep{pal2022medmcqalargescalemultisubject}. All benchmark datasets are obtained from the Hugging Face repository. Comprehensive descriptions of each evaluation dataset are provided in Appendix~\ref{sec:app_dataset}.

To assess the calibration performance of \methodname{}, we measure four established metrics: Expected Calibration Error ({ECE})\citep{naeini2015obtaining}, Maximum Calibration Error ({MCE}) \citep{naeini2015obtaining}, Adaptive Calibration Error ({ACE}) \citep{nixon2019measuring} and {Brier Score} \citep{brier1950verification}. Additional details are provided in Appendix \ref{app:details}.

\noindent\textbf{Baselines.} 
We compare our method with several post-hoc calibration techniques. Our unsupervised baselines include {DACA} \citep{luo2025your}, which aligns the pre-trained model on agreement samples; a hidden-state-based approach, Internal Consistency ({IC}) \citep{xie2024calibrating1}, which measures the ratio of consistency between each layer’s predictions and the final layer’s output; and two prompt-based methods: {CAPE} \citep{jiang2023calibrating}, which reduces bias by reordering answer choices, and {Elicitation} \citep{tian2023just}, which prompts the model to state its confidence. We also report results for the uncalibrated \textit{Vanilla} model and use supervised {Temperature Scaling (TS)} \citep{guo2017calibration} as an oracle. More details of baselines are presented in Appendix \ref{sec:baseline_app}.

\subsection{Main Results}
\label{sec:main-results}

\noindent\textbf{\methodname{} consistently achieves state-of-the-art results.}
\methodname{} demonstrates superior performance across all evaluated models and metrics, establishing a new state-of-the-art for unsupervised LLM calibration by outperforming all other unsupervised baselines, as shown in Table~\ref{tab:main_results}. For instance, on MMLU with the Llama-3.1-8B, our method achieves an ECE of just 2.871\%, a significant reduction compared to the 7.811\% of the strongest unsupervised baseline, DACA, and the 10.806\% of the uncalibrated model.  Notably, our framework's performance can significantly outperform the hidden-state-based approach IC and closely approach that of the supervised TS oracle. This indicates that our method that tackles both output drift and process drift in a dual alignment manner, can effectively address the complex dynamics of miscalibration while reducing human annotation costs. We also present the reliability diagrams visualization in Appendix~\ref{app:vis}.

\begin{table}
    \centering
    \small
    \setlength{\tabcolsep}{4pt}      
    \renewcommand{\arraystretch}{1.05} 

    \begin{tabular}{@{}llcc@{}}
        \toprule
        \textbf{Size} & \textbf{Method} & \textbf{ECE} ($\downarrow$) & \textbf{MCE} ($\downarrow$) \\
        \midrule
        \multirow{3}{*}{7B} 
        & Vanilla & $20.666_{\pm0.382}$ & $38.647_{\pm1.219}$\\
        & DACA & $10.312_{\pm0.502}$ & $16.884_{\pm0.954}$ \\
        & \textbf{\methodname{}} & $\mathbf{9.406_{\pm0.577}}$ & $\mathbf{15.256_{\pm0.993}}$ \\
        \midrule
        \multirow{3}{*}{14B} 
        & Vanilla & $23.842_{\pm0.336}$ & $58.230_{\pm8.103}$ \\
        & DACA & $5.146_{\pm0.340}$ & $\mathbf{8.867_{\pm0.590}}$\\
        & \textbf{\methodname{}} & $\mathbf{2.423_{\pm0.070}}$ & $11.241_{\pm2.918}$ \\
        \midrule
        \multirow{3}{*}{32B} 
        & Vanilla & $11.338_{\pm0.065}$ & $23.522_{\pm5.214}$ \\
        & DACA & $10.958_{\pm0.670}$ & $17.312_{\pm1.082}$ \\
        & \textbf{\methodname{}} & $\mathbf{9.203_{\pm0.055}}$ & $\mathbf{15.723_{\pm0.332}}$ \\
        \bottomrule
    \end{tabular}
    \caption{\small \textbf{Evaluation of \methodname{} with different model sizes.}
    We experiment with Qwen2.5 series of different model sizes.}
    \label{tab:model_sizes}
    \vspace{-1.5em}
\end{table}

\noindent\textbf{\methodname{} is effective across different model architectures and sizes.}
To validate the scalability and generalizability of our method, we conduct experiments across different model architectures (Qwen2.5-14B and Gemma-3-27B) in Table~\ref{tab:main_results}, and the Qwen-2.5 model series with varying sizes in Table~\ref{tab:model_sizes}. The results demonstrate that our method can maintain its effectiveness as model architecture varies and model size increases from 7B to 32B parameters. In all configurations, our method consistently outperforms both the uncalibrated model and the DACA baseline. This consistent performance advantage across different model scenarios highlights that \methodname{} is not tailored to a specific model but is a general solution that can be applied practically and flexibly.

\subsection{Ablation Study}
\label{sec:ablation}
To validate the key components of our \methodname{} framework, we conduct a series of ablation studies on the MMLU benchmark using the Llama-3.1-8B model. We investigate the contributions of our dual-component loss function and our dynamic layer selection strategy.

\noindent\textbf{Ablation on loss components.} To validate our dual-component loss, we compare the full \methodname{} framework against three variants: one using only the confidence alignment loss ($\mathcal{L}_{\rm Conf}$) ("Conf Only"), one using only the process alignment loss ($\mathcal{L}_{\rm Process}$) ("Process Only"), and one that applies confidence alignment to agreement samples and process alignment to disagreement samples ("Simple Stratify"). As shown in Table~\ref{tab:loss_ablation}, the "Conf Only" variant is ineffective, performing worse than the DACA baseline. While the "Process Only" and "Simple Stratify" variants substantially reduce calibration error, our full \methodname{} framework—which dynamically integrates both losses—achieves the best overall performance. It significantly outperforms the ablated versions and approaches the results of the supervised TS oracle, confirming the necessity of our dual-component strategy for effective calibration.

\begin{table}
    \centering
    \small
    \setlength{\tabcolsep}{2pt}      
    \renewcommand{\arraystretch}{1.05} 

    \begin{tabular}{@{}llcc@{}}
        \toprule
        \textbf{Method} & \textbf{ECE (\%) $\downarrow$} & \textbf{MCE (\%) $\downarrow$} &  \\
        \midrule
        Vanilla & $10.806_{{\pm0.275}}$ & $18.602_{{\pm0.212}}$ & \\
        DACA & $7.811_{{\pm0.619}}$ & $13.824_{{\pm0.667}}$ &  \\
        \midrule
        \methodname{} (Conf Only) & $10.267_{{\pm0.925}}$ & $17.599_{{\pm1.145}}$ &  \\
        \methodname{} (Process Only) & $6.082_{{\pm1.982}}$ & $9.082_{{\pm3.011}}$ &  \\
        \methodname{} (Simple Stratify) & $5.547_{{\pm0.874}}$ & $7.725_{{\pm2.121}}$ &  \\
        \textbf{\methodname{} (Ours)} & $\bf{2.871_{{\pm0.308}}}$ & $\bf{5.587_{{\pm0.648}}}$ &  \\
        \midrule
        \rowcolor[gray]{0.9}
            TS$^\dagger$ (Oracle) & $1.526_{{\pm0.450}}$ & $4.790_{{\pm1.090}}$ &  \\
        \bottomrule
    \end{tabular}
    \vspace{-0.5em}

    \caption{\small\textbf{Ablation study on the loss components of \methodname{} using Llama-3.1-8B on the MMLU datasets.} Our full, dual alignment method significantly outperforms the ablated versions, highlighting the necessity of addressing both output and process drift.}
\label{tab:loss_ablation}
\vspace{-1.5em}
\end{table}


\noindent\textbf{Ablation on layer selection.} To validate our dynamic Peak Divergence Layer (PDL) selection strategy, we compare it against starting process alignment at fixed network depths ($L/4$, $L/2$, and $3L/4$). As shown in Table \ref{tab:layer_ablation}, our dynamic approach, which identifies the layer with the maximum JSD increase, yields substantially better calibration performance than any fixed-layer strategy. This result confirms that divergence is sample-dependent and that accurately identifying this layer on a per-sample basis is critical to the success of the \methodname{} framework.

\begin{table}[t]
    \centering
    \small
    \setlength{\tabcolsep}{4pt}      
    \renewcommand{\arraystretch}{1.05} 

    \begin{tabular}{@{}llcc@{}}
        \toprule
        \textbf{Method} & \textbf{ECE (\%) $\downarrow$} & \textbf{MCE (\%) $\downarrow$} &  \\
        \midrule
        Vanilla & $10.806_{{\pm0.275}}$ & $18.602_{{\pm0.212}}$ &  \\
        DACA & $7.811_{{\pm0.619}}$ & $13.824_{{\pm0.667}}$ & \\
        \midrule
        \methodname{} ($L/4$) & $4.716_{{\pm0.397}}$ & $9.089_{{\pm1.298}}$ &  \\
        \methodname{} ($L/2$) & $4.862_{{\pm0.363}}$ & $9.235_{{\pm0.874}}$ &  \\
        \methodname{} ($3L/4$) & $2.846_{{\pm0.460}}$ & $5.806_{{\pm0.845}}$ & \\
        \textbf{\methodname{} (Ours)} & $\bf{2.382_{{\pm0.619}}}$ & $\bf{4.928_{{\pm1.030}}}$ &  \\
        \midrule
        \rowcolor[gray]{0.9}
        TS$^\dagger$ (Oracle) & $1.526_{{\pm0.450}}$ & $4.790_{{\pm1.090}}$ &  \\
        \bottomrule
    \end{tabular}

\caption{\small\textbf{Ablation study on the PDL selection strategy of \methodname{} using Llama-3.1-8B on the MMLU datasets.} Our proposed method, which selects the layer with the maximum JSD increase, yields the best calibration performance.}
\label{tab:layer_ablation}
\vspace{-1em}
\end{table}

\section{Discussions}
\label{sec:discussion}
In this section, we explore the broader applicability and potential extensions of our proposed \methodname{} framework. We demonstrate its adaptability by showing its effectiveness on open-ended generation tasks, its successful generalization to specialized domains like medicine (see Appendix~\ref{app:med} for full results), and its compatibility with various post-training methodologies.


\paragraph{Can \methodname{} be used for open-ended tasks?}
While \methodname{} is designed for multiple-choice questions, it can be extended to open-ended generation via reformulation using the $p(\text{true})$ framework~\citep{kadavath2022language}. Specifically, the model generates a free-form answer and then self-evaluates it, enabling calibration without modifying the core method. As shown in Figure~\ref{fig:openqa}, \methodname{} consistently reduces ECE and MCE on TruthfulQA~\citep{lin2022truthfulqameasuringmodelsmimic}. This demonstrates that our framework successfully adapts to open-ended generation, outperforming the strong DACA baseline on both LLama-3.1-8B and Qwen2.5-14B models and proving its versatility beyond multiple-choice formats.


\paragraph{Applicability to other post-training methods.}
To demonstrate the general applicability of our \methodname{} framework, we evaluate its performance on models subjected to various popular post-training techniques. We evaluate \methodname{} on Qwen2.5-7B models trained with PPO~\citep{schulman2017proximalpolicyoptimizationalgorithms}, DPO~\citep{rafailov2023direct}, and GRPO~\citep{deepseekai2024deepseekv2strongeconomicalefficient}. As shown in Figure~\ref{fig:post-train}, \methodname{} consistently outperforms both the uncalibrated model and the DACA baseline across all settings, indicating that the proposed framework generalizes beyond instruction tuning to diverse post-training paradigms.

\section{Related Works}
\textbf{Post-training} refines LLMs after their initial pre-training on broad datasets \citep{tie2025survey, kumar2025llm}. This stage includes methods like full fine-tuning for task-specific adaptation \citep{yue2023disc, luo2023empirical}, Parameter-Efficient Fine-Tuning (PEFT) such as LoRA for resource-efficient specialization \citep{hu2022lora, gao2023llama, luong2024reft}, and reinforcement learning techniques like RLHF and DPO to align models with user preferences \citep{ouyang2022training, rafailov2023direct}. While creating versatile and aligned models, these post-training processes can introduce miscalibration. Our paper therefore investigates these effects and proposes a novel framework to calibrate Post-trained Language Models.

\noindent\textbf{Confidence calibration} aims to ensure a model's output confidence accurately reflects its correctness likelihood \citep{guo2017calibration}. However, studies show that post-training often leads to overconfident LLMs \citep{xiao2022uncertainty, chen2022close, liu2023litcab, jiang2023calibrating}. Current calibration approaches include eliciting verbalized confidence through prompting or fine-tuning \citep{lin2022teaching, tian2023just, yang2024verbalized, xie2024calibrating, leng2025tamingoverconfidencellmsreward, damani2025beyond, tao2025revisiting, li2025conftuner, zhou2025steerconf}, and estimating confidence from output logits \citep{shen2024thermometer, luo2025your, vejendla2025efficient}. Closest to our work, \cite{shen2024thermometer, xie2024calibrating} leverage hidden states for calibration. However, they fail to account for both the confidence / process drifts and alignment dynamics induced by post-training in one unified framework, which are central to our research.

\section{Conclusion}
We study overconfidence in post-trained LLMs and show that miscalibration arises from two mechanisms: output drift and process drift. We propose \methodname{}, an unsupervised post-hoc dual-alignment framework that corrects output drift via final-distribution matching and mitigates process drift by locating the Peak Divergence Layer and aligning subsequent Inferential Stability Entropy. \methodname{} adaptively balances these objectives using intermediate predictive divergence, learning a single temperature parameter without human labels. Experiments demonstrate state-of-the-art calibration across diverse LLMs and datasets. We hope this diagnosis and framework motivate further study of how post-training affects calibration.

\section*{Limitations}

Following literature~\cite{guo2017calibration,luo2025your}, our method performs post-hoc confidence calibration by learning a temperature conditioned on the predictions of a pretrained LLM. Future exploration on alternative model references (e.g., multimodal models) or training-based method is a promising research direction to LLM calibration.

\section*{Ethical Considerations}
Our work proposes a post-hoc confidence calibration method that does not modify model parameters or introduce new data or capabilities. However, improving calibration may increase user trust in model outputs that can still be incorrect; calibrated confidence should not be interpreted as a guarantee of correctness. We recommend using calibrated confidence alongside complementary safeguards such as verification, human oversight, and downstream safety checks, especially in high-stakes settings. We only use publicly available benchmark datasets and follow their original licenses and guidelines. These benchmarks do not contain personally identifiable information, and our experiments do not involve collecting or processing personal data; we also do not intentionally create or curate offensive content.



\bibliography{custom}

\newpage
\appendix

\addcontentsline{toc}{section}{Appendix}
\renewcommand{\thepart}{} 
\renewcommand{\partname}{} 
\part{Appendix} 


\section{Experimental Details}
\label{app:exp_details}

\subsection{Models Details}
\label{app:model_details}
We conduct our experiments across a diverse set of large language models, spanning various architectures and scales from prominent model families. Table~\ref{tab:model_details} provides a detailed overview of the specific pre-trained and post-trained versions used in this study.

\subsection{Datasets Details}
\label{sec:app_dataset}
We evaluate our method on three diverse benchmarks. MMLU~\citep{mmlu} is a widely-adopted benchmark for measuring massive multitask language understanding. MedMCQA~\citep{pal2022medmcqalargescalemultisubject} is a large-scale, multi-subject, multiple-choice question dataset designed for the medical domain. TruthfulQA~\citep{lin2022truthfulqameasuringmodelsmimic} is a benchmark used to measure a model's truthfulness and its ability to avoid generating falsehoods.

For all datasets, we divide the data into a 30\% subset for alignment training and a 70\% test set. 
All three datasets are publicly available on Hugging Face\footnote{%
  \url{https://huggingface.co/datasets/cais/mmlu}\newline
  \hspace*{1.8em}\url{https://huggingface.co/datasets/openlifescienceai/medmcqa}\newline
  \hspace*{1.8em}\url{https://huggingface.co/datasets/domenicrosati/TruthfulQA}%
}. 
For MMLU, we use the test split from all subjects, while for MedMCQA, we use the validation split.

\begin{table*}[b]
\centering
\renewcommand\arraystretch{1.2}
\setlength{\tabcolsep}{6mm}
\resizebox{\textwidth}{!}{
\begin{tabular}{lcc}
\toprule
\textbf{Model Family} & \textbf{Model Type} & \textbf{HuggingFace Path} \\
\midrule
\multirow{2}{*}{Llama-3.1 Family} & Pre-trained Model & \texttt{meta-llama/Llama-3.1-8B} \\
 & Post-trained Model & \texttt{meta-llama/Llama-3.1-8B-Instruct} \\
\midrule
\multirow{2}{*}{Qwen-2.5 Family} & Pre-trained Model & \texttt{Qwen/Qwen2.5-14B} \\
 & Post-trained Model & \texttt{Qwen/Qwen2.5-14B-Instruct} \\
\midrule
\multirow{2}{*}{Gemma-3 Family} & Pre-trained Model & \texttt{google/gemma-3-27b-pt} \\
 & Post-trained Model & \texttt{google/gemma-3-27b-it} \\
\bottomrule
\end{tabular}
}
\caption{\small An overview of models used in our experiments, detailing the pre-trained and post-trained versions and their respective Hugging Face paths for each family.}
\label{tab:model_details}
\end{table*}

\begin{figure*}[b]
  \centering
  \begin{subfigure}[t]{0.49\textwidth}
    \centering
    \includegraphics[width=1\linewidth, page=1]{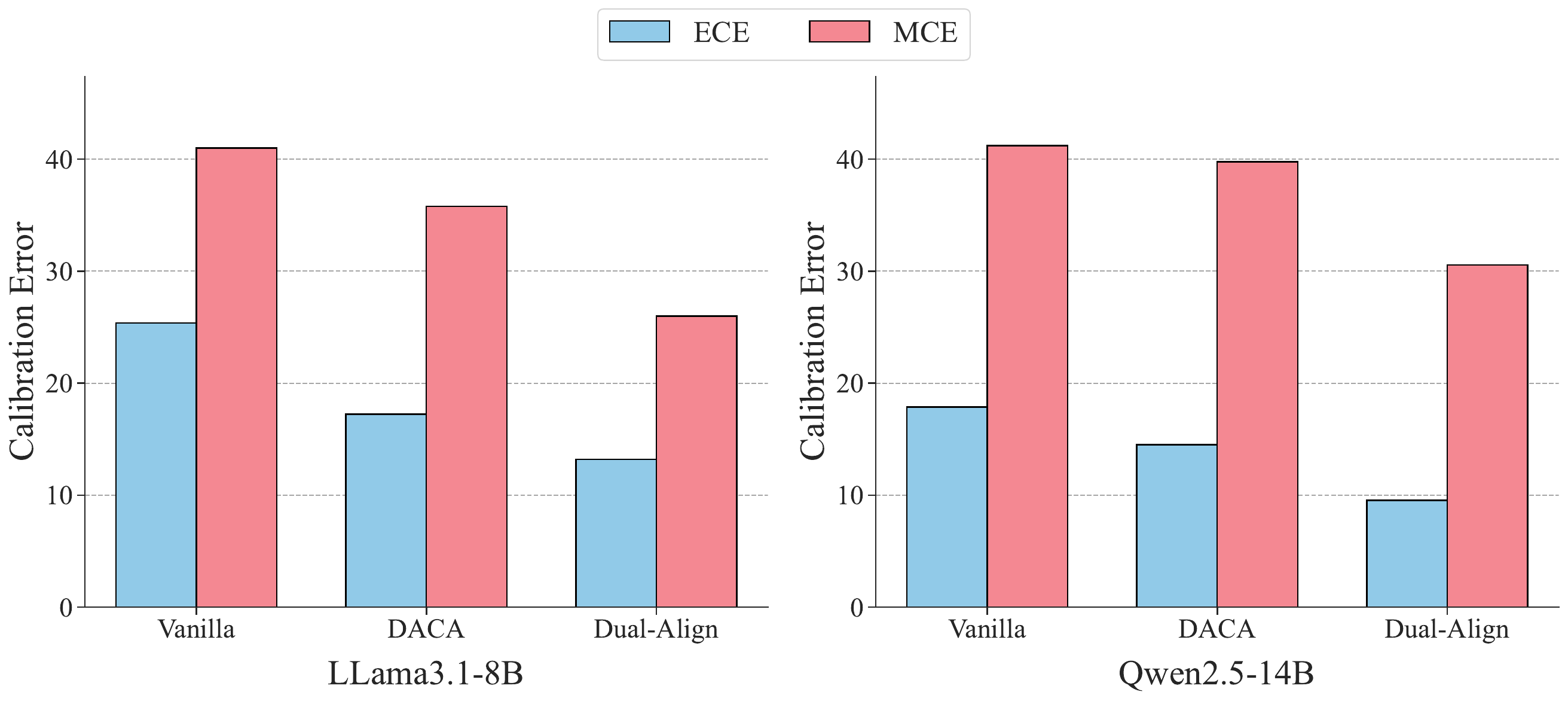}
    \caption{}
    \label{fig:openqa}
  \end{subfigure}
  \hfill
  \vspace{1em}
  \begin{subfigure}[t]{0.49\textwidth}
    \centering
\includegraphics[width=1\linewidth, page=1]{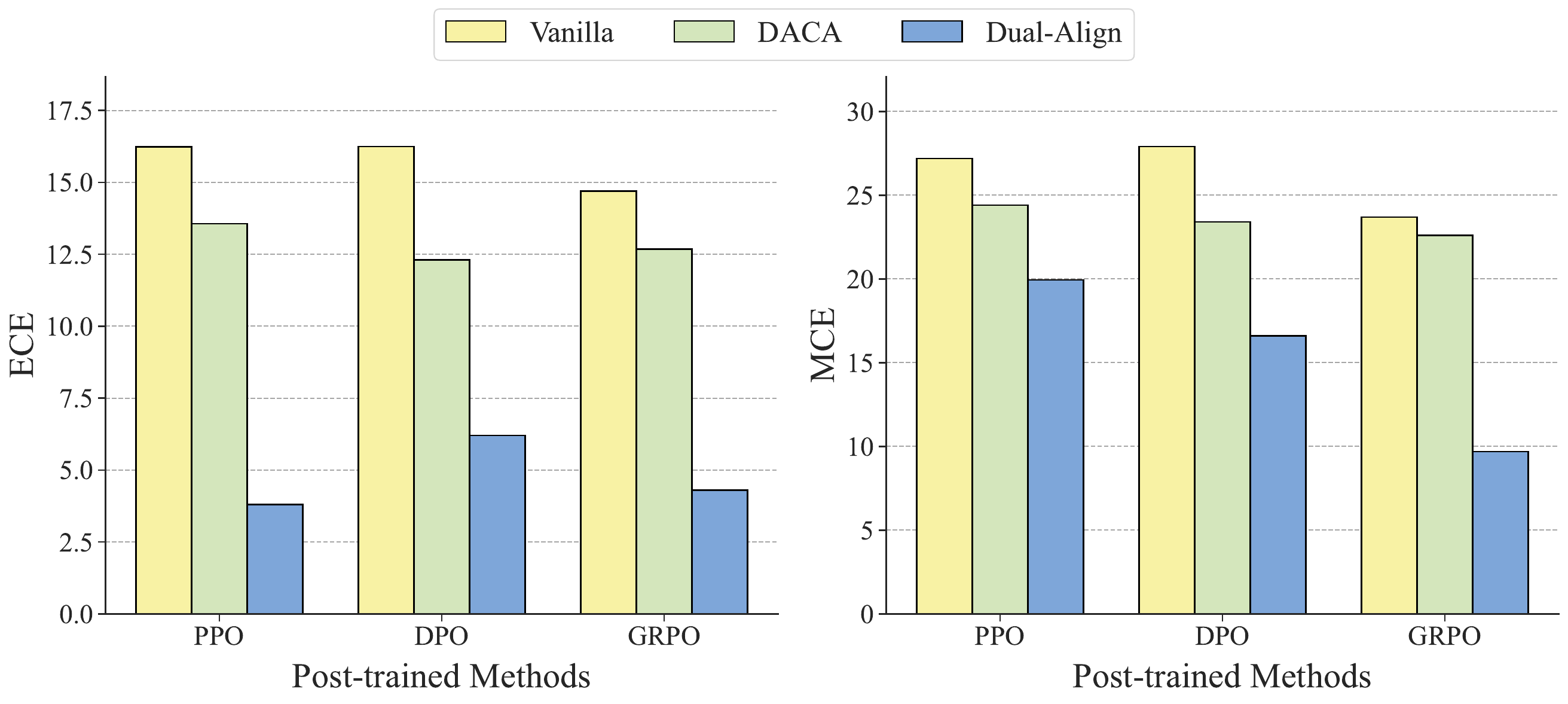}
    \caption{}
    \label{fig:post-train}
  \end{subfigure}
  \label{fig:}
  \caption{(a) \small{Applicability to open-ended question answering.} We evaluate LLama3.1 and Qwen2.5-14B on TruthfulQA dataset. (b) \small{Applicability to different post-training methods.} Apart from instruction-tuning, we consider PPO, DPO and GRPO training on Qwen2.5-7B.}
\end{figure*}

\subsection{Implementation Details}
\label{app:details}
All results are reported as mean $\pm$ standard deviation from three independent runs with different random seeds. All post-hoc methods requiring optimization—including our supervised oracle (Temperature Scaling) and the unsupervised baselines (DACA, \methodname{})—are trained using the Adam optimizer with a fixed learning rate of \texttt{0.05} for \texttt{300} epochs. For the unsupervised methods, we use a batch size of \texttt{128}. Finally, all bin-based calibration metrics (ECE, MCE, ACE) are computed using a default of 10 bins as specified in our evaluation script. For prompt templates used for evaluation, we present the details in Appendix~\ref{app:prompts}.

\subsection{Baseline Details}
\label{sec:baseline_app}
For prompt-based baselines, including CAPE \citep{jiang2023calibrating}: a prompt-based method that calibrates next-token probabilities by permuting option order to mitigate LLM biases, Elicitation \citep{tian2023just}: estimates confidence by prompting the model to generate verbalized probabilities. Unsupervised baseline DACA \citep{luo2025your} directly aligns the confidence of PoLMs to PLMs on the agreement samples. Internal Consistency (IC) \citep{xie2024calibrating1} measures the ratio of consistency between each layer’s predictions (mapped to the final vocabulary) and the final layer’s output. It is worth noting that the original IC leverages internal consistency within the model's reasoning process. Here, we ignore reasoning and directly generate the final answer for calculation. Since Elicitation and IC can only output confidence for prediction classes, we do not calculate the Brier Score.

\section{Comparison of Reliability Diagrams: PLM vs. PoLM}
\begin{figure*}[t]
    \vspace{-1em}
    \centering
    \begin{minipage}{0.24\linewidth}
        \centering
        \includegraphics[width=\linewidth]{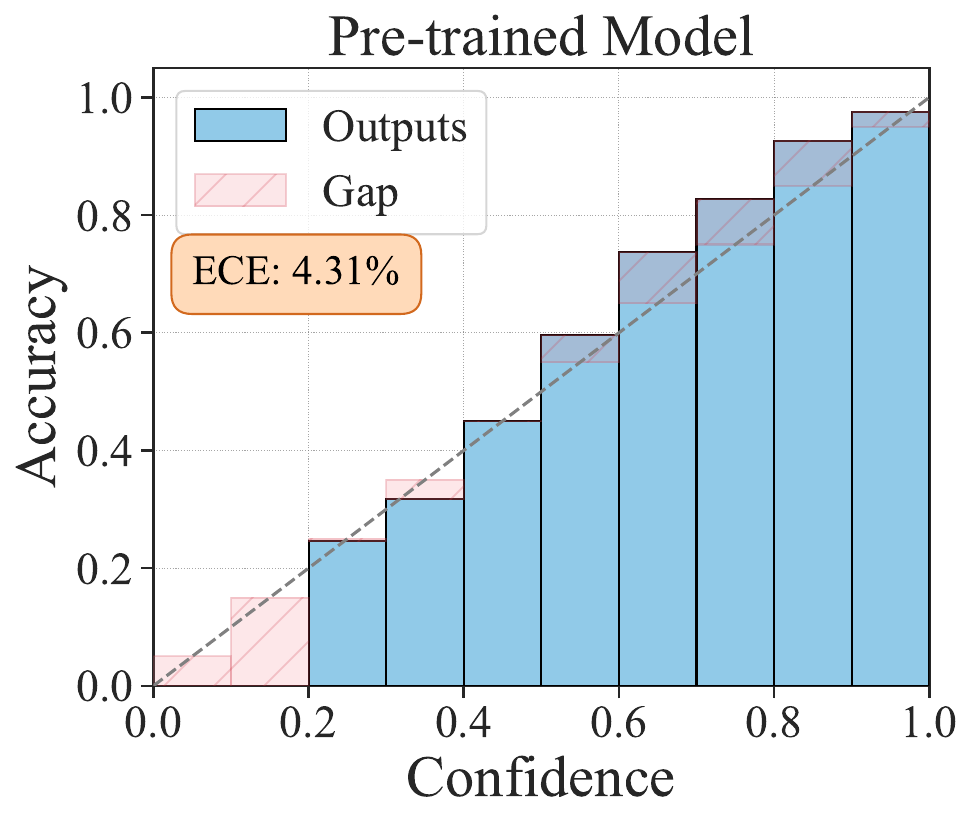}
    \end{minipage}\hfill
    \begin{minipage}{0.24\linewidth}
        \centering
        \includegraphics[width=\linewidth]{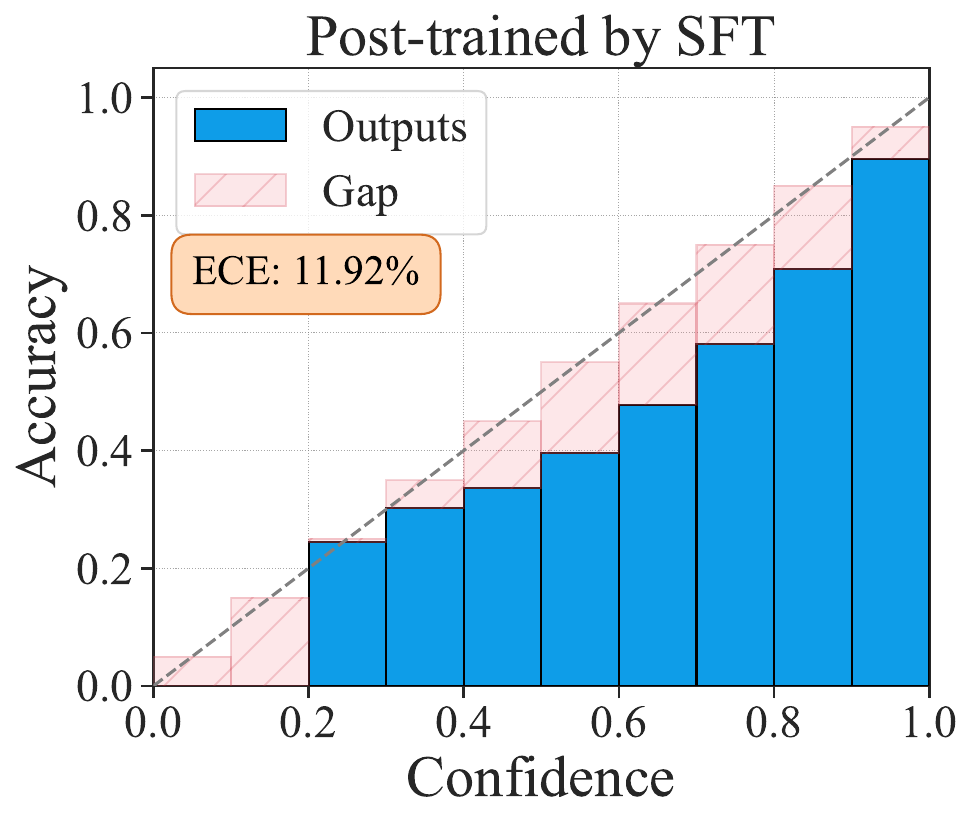}
    \end{minipage}\hfill
    \begin{minipage}{0.24\linewidth}
        \centering
        \includegraphics[width=\linewidth]{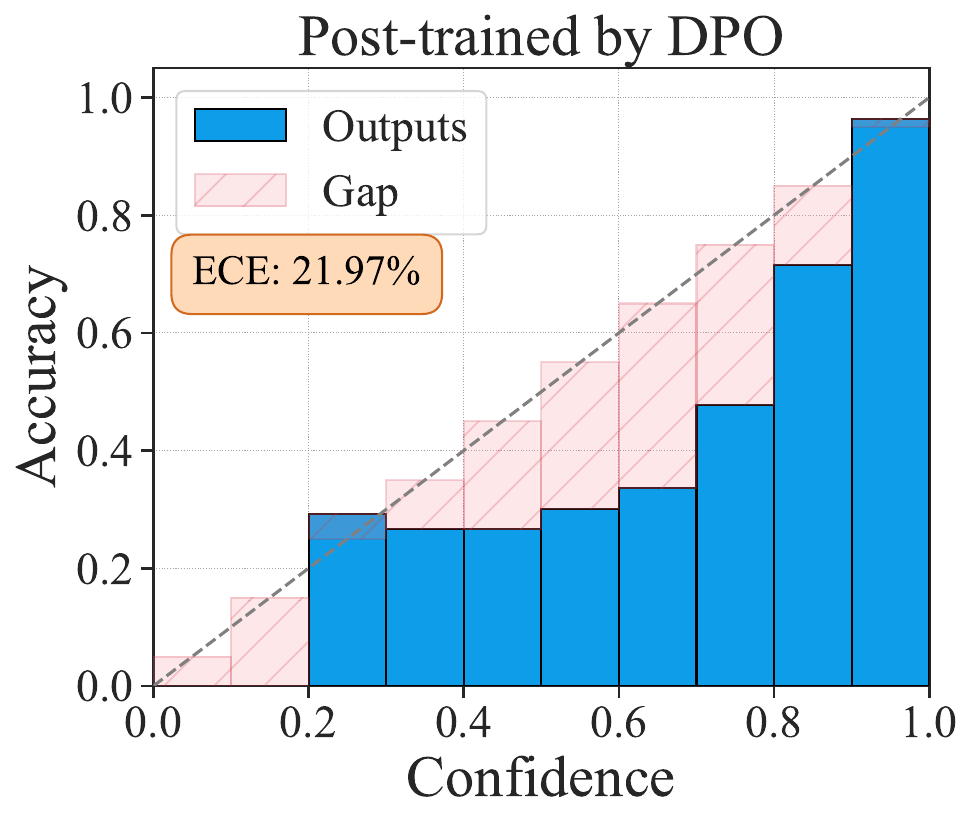}
    \end{minipage}\hfill
    \begin{minipage}{0.24\linewidth}
        \centering
        \includegraphics[width=\linewidth]{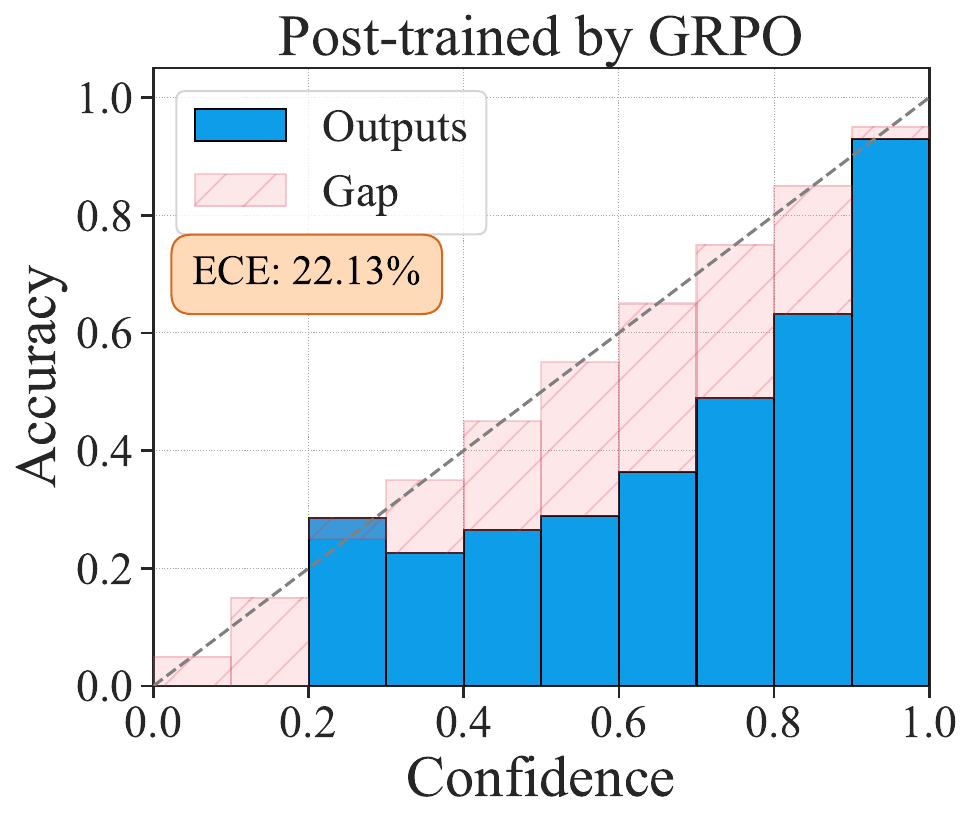}
    \end{minipage}
    \vspace{-0.8em}
        \caption{\small {Reliability diagrams on MMLU comparing a PLM with PoLMs obtained through various post-training methods.} The pre-trained model is Llama-3.1-8B-Base and the post-training techniques include Supervised Fine-tuning (SFT), Direct Preference Optimization (DPO) and Group Relative Policy Optimization (GRPO). }
    \label{fig:reliability}
    \vspace{-1em}
\end{figure*}

In this section, we present reliability diagrams for Llama-3.1-8B and its various post-trained versions on MMLU in Figure \ref{fig:reliability}. The results show that the pre-trained model is well-calibrated, while the post-trained versions exhibit significant overconfidence.

\section{More experiment results}
In this section, we present the results in Section \ref{sec:discussion} about entension to the open-ended question answering and other post-training methods, as shown in Figure \ref{fig:openqa} and Figure \ref{fig:post-train}.

\section{Evaluation on Other Domains}
\label{app:med}
In our main experiments, we conduct our evaluation on MMLU~\citep{mmlu} dataset. To further validate the generalizability of our method, we also present results on the MedMCQA~\citep{pal2022medmcqalargescalemultisubject} dataset, which is from the medical domain. All experimental settings are kept consistent with our main evaluation to ensure a fair comparison. The comprehensive results are shown in Table~\ref{tab:medmcqa}.

\section{Effect of Different Prompts}
\label{app:prompts}
To test our framework's robustness against prompt sensitivity, we evaluated four prompt templates (Figure \ref{fig:prompt_variations_english}). The results in Table \ref{tab:abl_prompt} confirm that \methodname{} consistently outperforms the baselines across all variants, demonstrating its effectiveness is not contingent on specific prompt phrasing and is robust to minor instructional changes.

\section{Reliability Diagrams of Different Baselines}
\label{app:vis}

This section provides reliability diagrams to visually assess calibration performance across our experiments. These plots show model accuracy versus confidence, with perfect calibration represented by the diagonal line. The following figures (Figure~\ref{start} to Figure~\ref{end}) present these diagrams for the uncalibrated (Vanilla) model, the DACA baseline, our \methodname{} framework, and the supervised Temperature Scaling (TS) oracle. These visualizations visually confirm the quantitative results from the main paper, clearly illustrating that \methodname{} significantly reduces the overconfidence of post-trained models and achieves a calibration profile that closely approaches the supervised oracle.

\section{Theory for Process Alignment}
\label{app:theory}

This appendix formalizes Proposition~\ref{prop:process_align_calibration}.
We analyze a smooth calibration surrogate built on a temperature-scaled logit-gap confidence proxy, and show that the resulting calibration error can be controlled by the ISE mismatch to a PLM reference, up to a PLM-dependent constant.

\subsection{Setup and Notation}
Let $f$ denote the post-trained language model (PoLM) and $g$ denote the pre-trained language model (PLM).
For an input $\bm{x}$ with multiple-choice option set $\mathcal{Y}$, let $\hat y=\arg\max_{i\in\mathcal Y}p_i^{L}(\bm{x})$ be the PoLM prediction at the final layer.
Define the (final-layer) logit gap
\begin{equation}
\Delta(\bm{x})\;:=\; z^{L}_{\hat y}(\bm{x})\;-\;\log\!\sum_{j\in\mathcal{Y}\setminus\{\hat y\}}\exp\!\big(z^{L}_{j}(\bm{x})\big).
\end{equation}
Given a temperature $\tau>0$ and the sigmoid function $\sigma(\cdot)$, define the confidence proxy
\begin{equation}
c_\tau(\bm{x})\;:=\;\sigma\!\big(\Delta(\bm{x})/\tau\big),
\end{equation}
and the squared calibration surrogate
\begin{equation}
\mathcal{E}_f(\tau)\;:=\;\mathbb{E}_{\bm{x}}\Big[\big(c_\tau(\bm{x})-\Pr(Y{=}y\mid \bm{x})\big)^2\Big],
\end{equation}
where $y$ is the ground-truth answer and $\Pr(Y{=}y\mid \bm{x})$ denotes the true correctness likelihood.

\paragraph{Inferential Stability Entropy (ISE).}
Let $l^*(\bm{x})$ be the Peak Divergence Layer (PDL) defined in the main paper.
Let $v_f^l(\bm{x})$ denote the PoLM logit of its predicted option at layer $l$ (via LogitLens), and define a distribution over layers
\begin{align}
q^l_f(\bm{x},\tau)\;:=\;\frac{\exp\!\big(v_f^l(\bm{x})/\tau\big)}{\sum_{j=l^*(\bm{x})}^{L}\exp\!\big(v_f^j(\bm{x})/\tau\big)},
\end{align}
where $l\in\{l^*(\bm{x}),\dots,L\}$.
The PoLM ISE under temperature $\tau$ is
\begin{equation}
\mathrm{ISE}_f(\bm{x},\tau)\;:=\;-\sum_{l=l^*(\bm{x})}^{L} q^l_f(\bm{x},\tau)\log q^l_f(\bm{x},\tau).
\end{equation}
We define $\mathrm{ISE}_g(\bm{x})$ analogously for the PLM (without temperature scaling, or with $\tau\!=\!1$).

\subsection{Assumptions}
We list mild regularity assumptions sufficient for the bound.

\paragraph{A1 (Margin-link model with sample-dependent scale).}
There exists a (possibly unknown) sample-dependent scale $\kappa(\bm{x})>0$ such that the true correctness probability can be written as
\begin{equation}
\Pr(Y{=}y\mid \bm{x})\;=\;\sigma\!\Big(\Delta(\bm{x})/\kappa(\bm{x})\Big).
\label{eq:assump_margin_scale}
\end{equation}
This captures the view that miscalibration arises from using a global $\tau$ to approximate a heterogeneous scale $\kappa(\bm{x})$.

\paragraph{A2 (Stability scale is controlled by PLM stability signal).}
There exists a (possibly unknown) function $\psi$ such that $\kappa(\bm{x})=\psi(\mathrm{ISE}_g(\bm{x}))$.
This formalizes that the PLM's stability profile provides a reference signal for the latent correctness scale.

\paragraph{A3 (ISE sensitivity to inverse temperature).}
There exists a constant $\lambda(\bm{x})>0$ such that for all $\tau$ in the relevant range,
\begin{equation}
\big|\mathrm{ISE}_f(\bm{x},\tau)-\mathrm{ISE}_f(\bm{x},\kappa(\bm{x}))\big|
\;\ge\;
\lambda(\bm{x})\Big|\tfrac{1}{\tau}-\tfrac{1}{\kappa(\bm{x})}\Big|.
\label{eq:assump_sensitivity}
\end{equation}
Intuitively, this excludes degenerate cases where post-PDL layer logits are nearly constant across layers, in which ISE barely changes with $\tau$.

\paragraph{A4 (Boundedness).}
Assume $|\Delta(\bm{x})|\le M$ and $\kappa(\bm{x})\in[\kappa_{\min},\kappa_{\max}]$ for constants $M>0$ and $0<\kappa_{\min}\le\kappa_{\max}$.

\subsection{A Helpful Inequality}
We use that the logistic function has bounded derivative:
\begin{equation}
\sup_{t\in\mathbb{R}}|\sigma'(t)|\;\le\;\tfrac{1}{4}.
\label{eq:sigma_deriv}
\end{equation}

\subsection{Main Result}
\begin{theorem}[Formal version of Proposition~\ref{prop:process_align_calibration}]
\label{thm:process_align_calibration_formal}
Under Assumptions A1--A4, for any $\tau>0$,
\begin{equation}
\mathcal{E}_f(\tau)
\le
\mathbb{E}_{\bm{x}}\left[w(\bm{x})\big(\mathrm{ISE}_f(\bm{x},\tau)-\mathrm{ISE}_g(\bm{x})\big)^2\right]
+C_g,
\label{eq:thm_bound}
\end{equation}
where one valid choice is $w(\bm{x})=\frac{\Delta(\bm{x})^2}{16\,\lambda(\bm{x})^2}$, and
\begin{equation}
C_g:=\mathbb{E}_{\bm{x}}\left[w(\bm{x})\big(\mathrm{ISE}_f(\bm{x},\kappa(\bm{x}))-\mathrm{ISE}_g(\bm{x})\big)^2\right]
\ge0
\label{eq:Cg_def}
\end{equation}
is a PLM-dependent constant that does not depend on $\tau$.
\end{theorem}

\subsection{Proof of Theorem~\ref{thm:process_align_calibration_formal}}
\begin{proof}
By Assumption A1 and the definition of $c_\tau(\bm{x})$,
\[
c_\tau(\bm{x})-\Pr(Y{=}y\mid \bm{x})
=
\sigma\!\Big(\tfrac{\Delta(\bm{x})}{\tau}\Big)-\sigma\!\Big(\tfrac{\Delta(\bm{x})}{\kappa(\bm{x})}\Big).
\]
By the mean value theorem and~\eqref{eq:sigma_deriv},
\begin{equation}
\big|c_\tau(\bm{x})-\Pr(Y{=}y\mid \bm{x})\big|
\;\le\;
\tfrac{1}{4}\,|\Delta(\bm{x})|\cdot\Big|\tfrac{1}{\tau}-\tfrac{1}{\kappa(\bm{x})}\Big|.
\label{eq:mvt_step}
\end{equation}
Squaring~\eqref{eq:mvt_step} and taking expectation yields
\begin{equation}
\mathcal{E}_f(\tau)
\;\le\;
\mathbb{E}_{\bm{x}}\!\left[\tfrac{\Delta(\bm{x})^2}{16}\Big|\tfrac{1}{\tau}-\tfrac{1}{\kappa(\bm{x})}\Big|^2\right].
\label{eq:E_step1}
\end{equation}
Next, apply Assumption A3:
\[
\Big|\tfrac{1}{\tau}-\tfrac{1}{\kappa(\bm{x})}\Big|
\;\le\;
\lambda(\bm{x})^{-1}\big|\mathrm{ISE}_f(\bm{x},\tau)-\mathrm{ISE}_f(\bm{x},\kappa(\bm{x}))\big|.
\]
Plugging this into~\eqref{eq:E_step1} gives
\begin{equation}
\mathcal{E}_f(\tau)
\;\le\;
\mathbb{E}_{\bm{x}}\!\left[
\underbrace{\tfrac{\Delta(\bm{x})^2}{16\lambda(\bm{x})^2}}_{w(\bm{x})}
\big(\mathrm{ISE}_f(\bm{x},\tau)-\mathrm{ISE}_f(\bm{x},\kappa(\bm{x}))\big)^2
\right].
\label{eq:E_step2}
\end{equation}
Finally, add and subtract $\mathrm{ISE}_g(\bm{x})$ and use $(a-b)^2\le (a-c)^2+(c-b)^2$ (or the looser $2$-$2$ inequality):
\begin{equation}
\begin{aligned}
&\big(\mathrm{ISE}_f(\bm{x},\tau)-\mathrm{ISE}_f(\bm{x},\kappa(\bm{x}))\big)^2
\; \\ &\le\;
\big(\mathrm{ISE}_f(\bm{x},\tau)-\mathrm{ISE}_g(\bm{x})\big)^2
 \\ & +
\big(\mathrm{ISE}_f(\bm{x},\kappa(\bm{x}))-\mathrm{ISE}_g(\bm{x})\big)^2.
\end{aligned}
\end{equation}

Substituting into~\eqref{eq:E_step2} yields~\eqref{eq:thm_bound} with $C_g$ defined in~\eqref{eq:Cg_def}.
\end{proof}

\subsection{Remarks}
\paragraph{Why $C_g$ is PLM-dependent and benign.}
The constant $C_g$ does not depend on $\tau$ and quantifies how well the PLM stability signal $\mathrm{ISE}_g(\bm{x})$ matches the ``ideal'' PoLM stability $\mathrm{ISE}_f(\bm{x},\kappa(\bm{x}))$ associated with the true scale $\kappa(\bm{x})$.
When the PLM serves as a reliable stability reference, $C_g$ is small, and minimizing the ISE mismatch term yields a small upper bound on $\mathcal{E}_f(\tau)$.

\paragraph{On Assumption A1.}
Assumption~\eqref{eq:assump_margin_scale} is a standard smooth-link modeling choice for analysis; it formalizes that correctness depends on the final-layer margin with a sample-dependent scale (capturing heterogeneity induced by post-training).
Our empirical results support that aligning stability (ISE) improves calibration in practice.

\begin{table*}[t]
\centering
\renewcommand\arraystretch{1.4} 
\setlength{\tabcolsep}{4mm}     
\resizebox{1\textwidth}{!}{   
\begin{adjustbox}{max width=\textwidth}
\begin{tabular}{cl*{4}{c}}
  \toprule
  \multirow{2.5}*{\textbf{Models}} & \multirow{2.5}*{\textbf{Methods}} & \multicolumn{4}{c}{\textbf{Evaluation Metrics}} \\
  \cmidrule(lr){3-6}
    & & \textbf{ECE (\%)} $\downarrow$ & \textbf{MCE (\%)} $\downarrow$ & \textbf{ACE (\%)} $\downarrow$ & \textbf{Brier Score} $\downarrow$ \\ 

    \midrule
    \multirow{4.5}*{\rotatebox{90}{\textbf{LLama3.1-8B}}} 
    & Vanilla & $16.919_{{\pm0.699}}$ & $27.511_{{\pm0.424}}$ & $15.679_{{\pm1.388}}$ & $0.564_{{\pm0.005}}$ \\
    & DACA  & $5.149_{{\pm0.350}}$ & $10.582_{{\pm0.521}}$ & $5.729_{{\pm0.374}}$ & $0.517_{{\pm0.003}}$ \\
    & \textbf{\methodname~(\textbf{Ours})} & $\bf{4.684_{{\pm0.171}}}$ & $\bf{8.881_{{\pm0.393}}}$ & $\bf{5.106_{{\pm0.432}}}$ & $\bf{0.516_{{\pm0.003}}}$ \\
    \cmidrule(lr){2-6}
     & TS$^\dagger$ (oracle) & $1.587_{{\pm0.545}}$ & $4.929_{{\pm2.491}}$ & $1.842_{{\pm0.444}}$ & $0.513_{{\pm0.003}}$ \\

    \midrule
    \multirow{4.5}*{\rotatebox{90}{\textbf{Qwen2.5-14B}}} 
    & Vanilla & $26.881_{{\pm0.631}}$ & $39.386_{{\pm0.109}}$ & $23.303_{{\pm0.471}}$ & $0.621_{{\pm0.010}}$ \\
    & DACA & $4.904_{{\pm0.433}}$ & $9.245_{{\pm0.270}}$ & $8.361_{{\pm0.442}}$ & $0.529_{{\pm0.005}}$ \\
    & \textbf{\methodname~(\textbf{Ours})} & $\bf{3.538_{{\pm0.924}}}$ & $\bf{7.507_{{\pm0.866}}}$ & $\bf{3.483_{{\pm0.359}}}$ & $\bf{0.489_{{\pm0.006}}}$ \\
    \cmidrule(lr){2-6}
    & TS$^\dagger$ (oracle) & $3.628_{{\pm0.408}}$ & $19.972_{{\pm8.798}}$ & $7.184_{{\pm0.950}}$ & $0.498_{{\pm0.006}}$ \\

  \midrule
  \multirow{4.5}*{\rotatebox{90}{\textbf{Gemma-3-27B}}} 
    & Vanilla & $37.084_{\pm0.058}$ & $49.348_{\pm14.837}$ & $34.293_{\pm4.081}$ & $0.748_{\pm0.001}$ \\
    & DACA & $26.872_{\pm0.238}$ & $38.685_{\pm1.628}$ & $24.443_{\pm0.497}$ & $0.628_{\pm0.003}$ \\
    & \textbf{\methodname~(\textbf{Ours})} & $\bf{12.940_{\pm0.176}}$ & $\bf{29.034_{\pm0.220}}$ & $\bf{14.765_{\pm0.292}}$ & $\bf{0.537_{\pm0.001}}$ \\
    \cmidrule(lr){2-6}
     & TS$^\dagger$ (oracle) & $6.917_{\pm0.278}$ & $28.561_{\pm0.187}$ & $9.317_{\pm0.297}$ & $0.519_{\pm0.002}$ \\

  \bottomrule
\end{tabular}
\end{adjustbox}
}
\caption{\small Performance comparison across different PoLMs and calibration methods on MedMCQA datasets. Lower values indicate better performance. Best results among unsupervised methods are shown in \textbf{bold}. "Vanilla" refers to uncalibrated PoLMs. $\dagger$ indicates calibration methods with access to labels.  Values are percentages averaged over 3 runs.}
\label{tab:medmcqa}
\end{table*}

\begin{figure*}[t]
\centering
\begin{tcolorbox}[
    enhanced,
    colback=gray!10,
    colframe=gray!80!black,
    fonttitle=\bfseries,
    title={Prompt Variations for Multiple-Choice Questions}
]
\textbf{Prompt Variant A (used in main experiments)}\par
Select the correct answer for each of the following questions. Respond with the letter only:\par
\texttt{[Question]\newline
A: [Option 1] B: [Option 2] C: [Option 3] D: [Option 4]\newline
Answer: }

\vspace{1.2em}

\textbf{Prompt Variant B}\par
The following are multiple-choice questions. Give ONLY the correct option, no other words or explanation:\par
\texttt{[Question]\newline
A: [Option 1] B: [Option 2] C: [Option 3] D: [Option 4]\newline
Answer: }

\vspace{1.2em}

\textbf{Prompt Variant C}\par
For the following multiple choice question, provide just the correct letter:\par
\texttt{[Question]\newline
A: [Option 1] B: [Option 2] C: [Option 3] D: [Option 4]\newline
Answer: }

\vspace{1.2em}

\textbf{Prompt Variant D}\par
Directly select the correct answer for the following multiple choice question without any explanations:\par
\texttt{[Question]\newline
A: [Option 1] B: [Option 2] C: [Option 3] D: [Option 4]\newline
Answer: }

\end{tcolorbox}
\vspace{-1em}
\caption{\small Four different prompt instructions for a multiple-choice question task.}
\label{fig:prompt_variations_english}
\end{figure*}

\begin{table*}[t]
\centering
\renewcommand\arraystretch{1.4} 
\setlength{\tabcolsep}{4mm}     
\resizebox{1\textwidth}{!}{  
\begin{tabular}{ll*{4}{c}}
  \toprule
  \multirow{2.5}*{\textbf{Prompt Type}} & \multirow{2.5}*{\textbf{Methods}} & \multicolumn{4}{c}{\textbf{Evaluation Metrics}} \\
  \cmidrule(lr){3-6}
    & & \textbf{ECE (\%)} $\downarrow$ & \textbf{MCE (\%)} $\downarrow$ & \textbf{ACE (\%)} $\downarrow$ & \textbf{Brier Score} $\downarrow$ \\
    \midrule
    \multirow{4.5}*{Prompt A}
    & Vanilla & $10.806_{{\pm0.275}}$ & $18.602_{{\pm0.212}}$ & $11.809_{{\pm0.652}}$ & $0.461_{{\pm0.005}}$ \\
    & DACA & $7.811_{{\pm0.619}}$ & $13.824_{{\pm0.667}}$ & $8.064_{{\pm0.544}}$ & $0.451_{{\pm0.004}}$ \\
    & \textbf{\methodname~(Ours)} & $\bf{2.871_{{\pm0.308}}}$ & $\bf{5.587_{{\pm0.648}}}$ & $\bf{3.222_{{\pm0.306}}}$ & $\bf{0.441_{{\pm0.004}}}$ \\
    \cmidrule(lr){2-6}
     & TS$^\dagger$ (oracle) & $1.526_{{\pm0.450}}$ & $4.790_{{\pm1.090}}$ & $1.985_{{\pm0.609}}$ & $0.441_{{\pm0.004}}$ \\

    \midrule
    \multirow{4.5}*{Prompt B}
    & Vanilla & $13.271_{{\pm0.375}}$ & $23.224_{{\pm0.708}}$ & $13.917_{{\pm0.638}}$ & $0.472_{{\pm0.006}}$ \\
    & DACA & $5.530_{{\pm0.627}}$ & $10.027_{{\pm1.251}}$ & $6.196_{{\pm0.558}}$ & $0.444_{{\pm0.003}}$ \\
    & \textbf{\methodname~(Ours)} & $\bf{1.441_{{\pm0.127}}}$ & $\bf{8.835_{{\pm0.301}}}$ & $\bf{2.278_{{\pm0.225}}}$ & $\bf{0.439_{{\pm0.004}}}$ \\
    \cmidrule(lr){2-6}
    & TS$^\dagger$ (oracle) & $1.641_{{\pm0.341}}$ & $8.820_{{\pm0.132}}$ & $2.488_{{\pm0.424}}$ & $0.439_{{\pm0.004}}$ \\

  \midrule
  \multirow{4.5}*{Prompt C}
    & Vanilla& $10.183_{{\pm0.254}}$ & $18.464_{{\pm1.361}}$ & $10.859_{{\pm0.587}}$ & $0.456_{{\pm0.005}}$ \\
    & DACA & $6.435_{{\pm0.710}}$ & $11.929_{{\pm0.842}}$ & $6.830_{{\pm0.785}}$ & $0.444_{{\pm0.004}}$ \\
    & \textbf{\methodname~(Ours)} & $\bf{3.364_{{\pm0.385}}}$ & $\bf{6.659_{{\pm0.829}}}$ & $\bf{3.994_{{\pm0.380}}}$ & $\bf{0.439_{{\pm0.004}}}$ \\
    \cmidrule(lr){2-6}
    & TS$^\dagger$ (oracle) & $1.387_{{\pm0.237}}$ & $6.954_{{\pm1.340}}$ & $2.143_{{\pm0.294}}$ & $0.437_{{\pm0.004}}$ \\

    \midrule
    \multirow{4.5}*{Prompt D}
    & Vanilla & $11.860_{{\pm0.281}}$ & $21.147_{{\pm1.020}}$ & $13.414_{{\pm0.451}}$ & $0.470_{{\pm0.004}}$ \\
    & DACA DACA & $5.074_{{\pm0.528}}$ & $9.856_{{\pm0.162}}$ & $5.729_{{\pm0.632}}$ & $0.450_{{\pm0.003}}$ \\
    & \textbf{\methodname~(Ours)} & $\bf{2.523_{{\pm0.410}}}$ & $\bf{6.792_{{\pm1.148}}}$ & $\bf{3.031_{{\pm0.087}}}$ & $\bf{0.445_{{\pm0.003}}}$ \\
    \cmidrule(lr){2-6}
    & TS$^\dagger$ (oracle) & $1.915_{{\pm0.084}}$ & $5.849_{{\pm3.020}}$ & $2.370_{{\pm0.449}}$ & $0.445_{{\pm0.003}}$ \\

   \bottomrule
\end{tabular}
}
\caption{\small Effects of different prompt instructions on calibration error using Llama3.1-8B on MMLU dataset.}
\label{tab:abl_prompt}
\end{table*}

\begin{figure*}
    \centering
    \begin{minipage}{0.24\linewidth}
        \centering        \includegraphics[width=\linewidth]{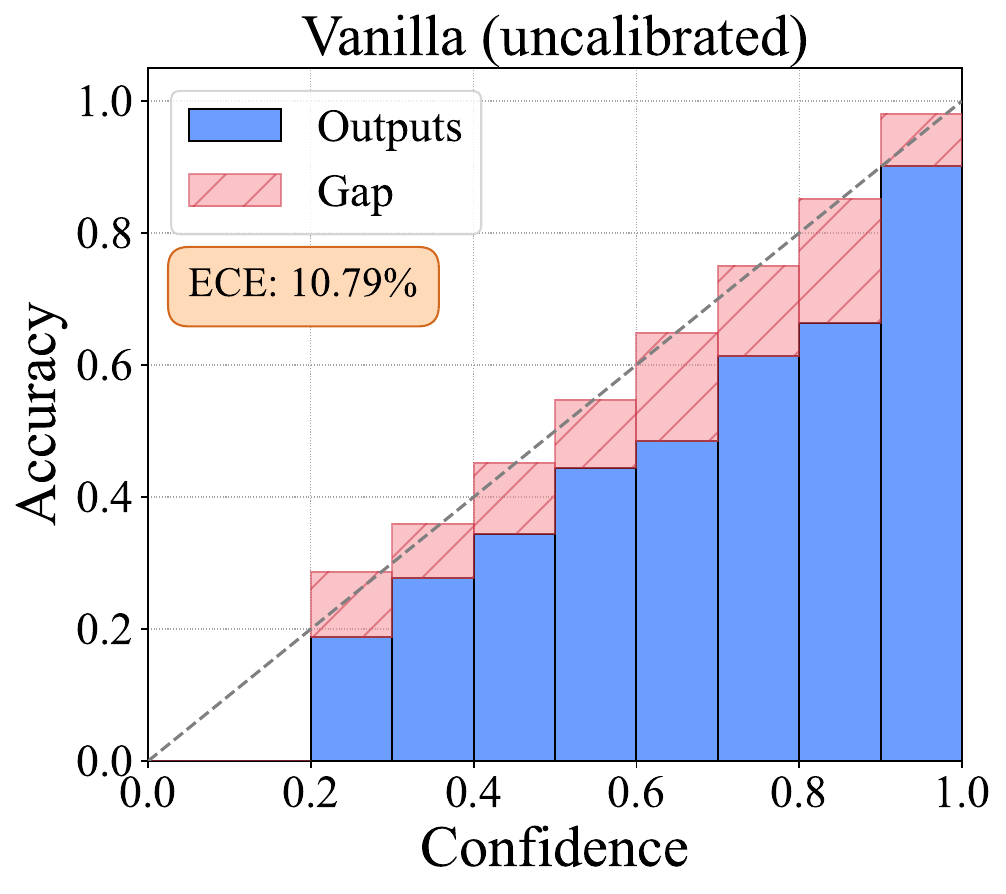}
    \end{minipage}\hfill
    \begin{minipage}{0.24\linewidth}
        \centering
        \includegraphics[width=\linewidth]{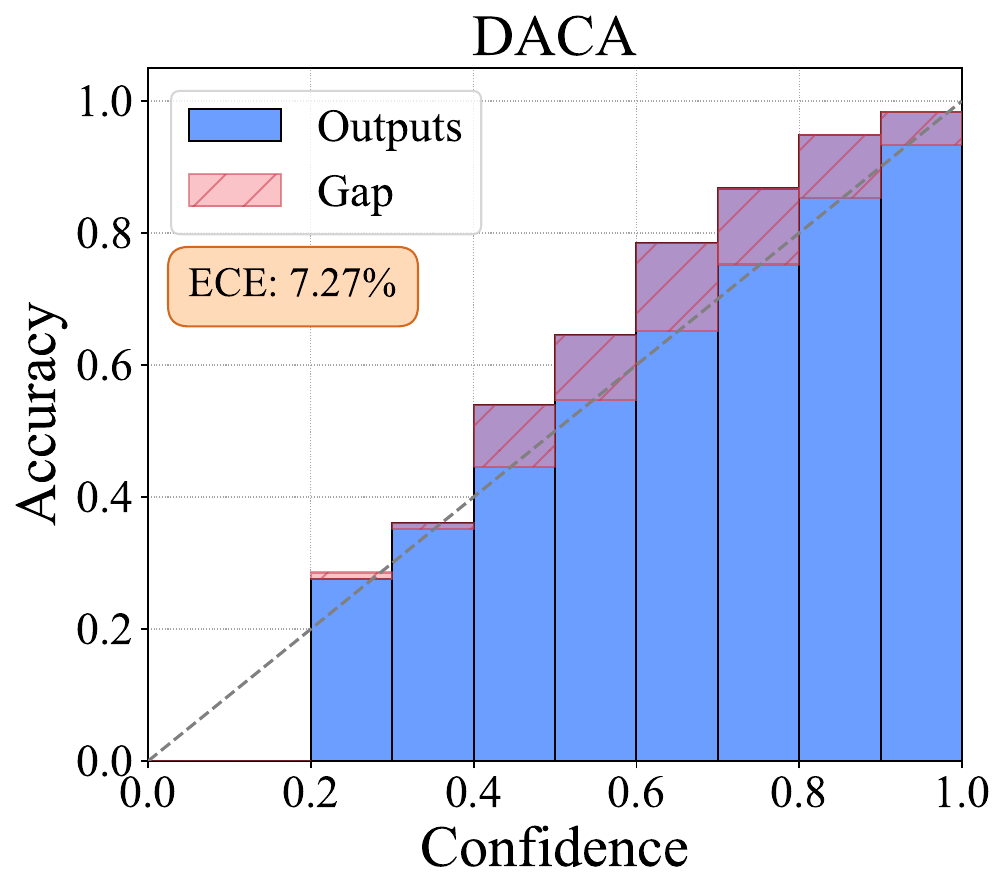}
    \end{minipage}\hfill
    \begin{minipage}{0.24\linewidth}
        \centering
        \includegraphics[width=\linewidth]{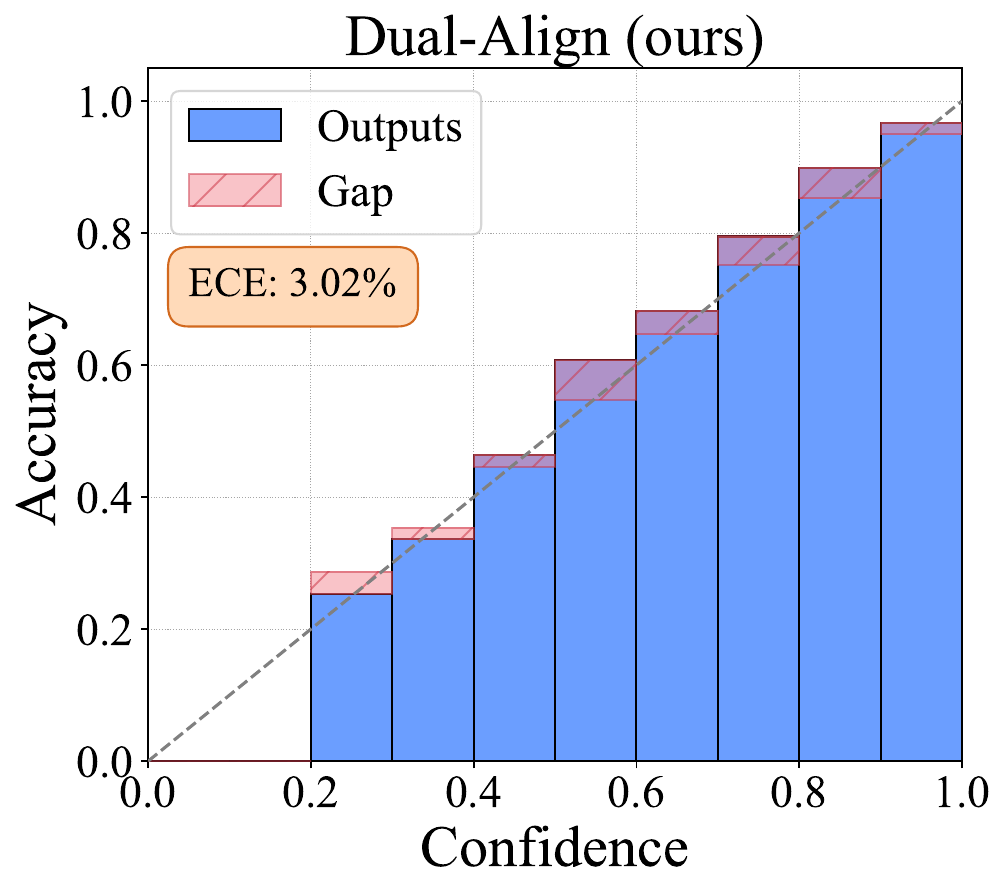}
    \end{minipage}\hfill
    \begin{minipage}{0.24\linewidth}
        \centering
        \includegraphics[width=\linewidth]{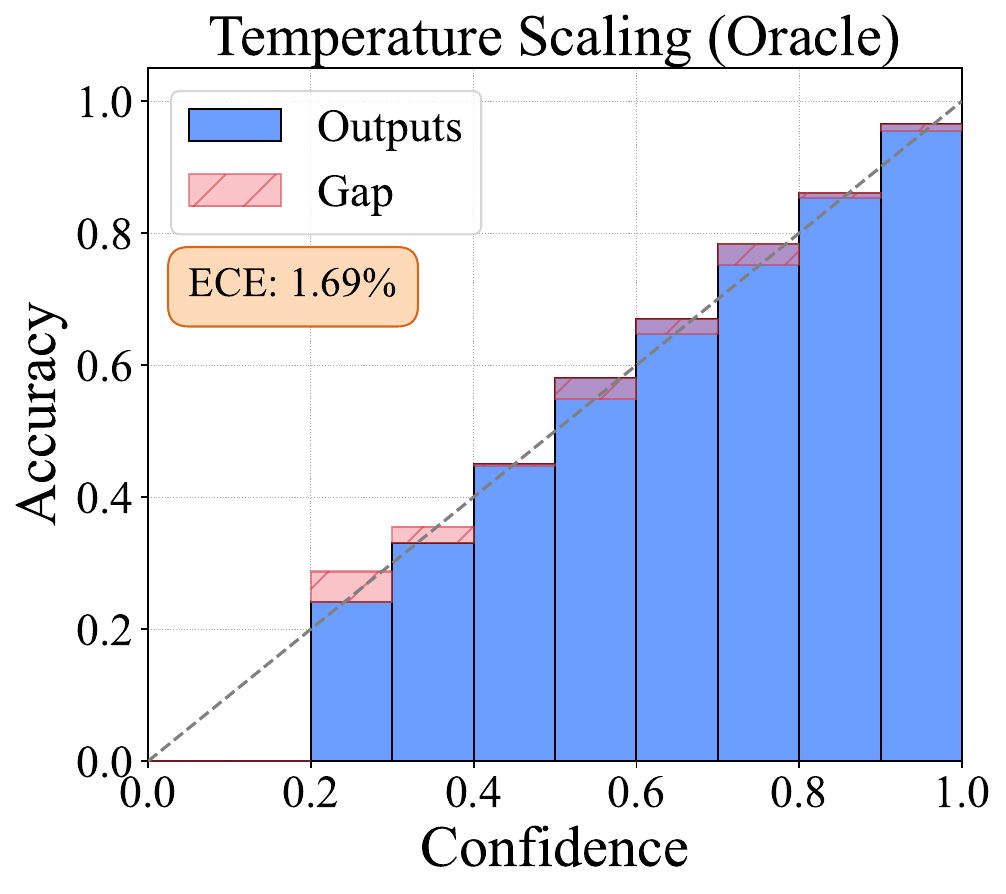}
    \end{minipage}
    \caption{\small Reliability diagrams of Llama3.1-8B-Instruct on MMLU dataset.}
    \label{start}
    \vspace{-1em}
\end{figure*}

\begin{figure*}[!h]
    \centering
    \begin{minipage}{0.24\linewidth}
        \centering
        \includegraphics[width=\linewidth]{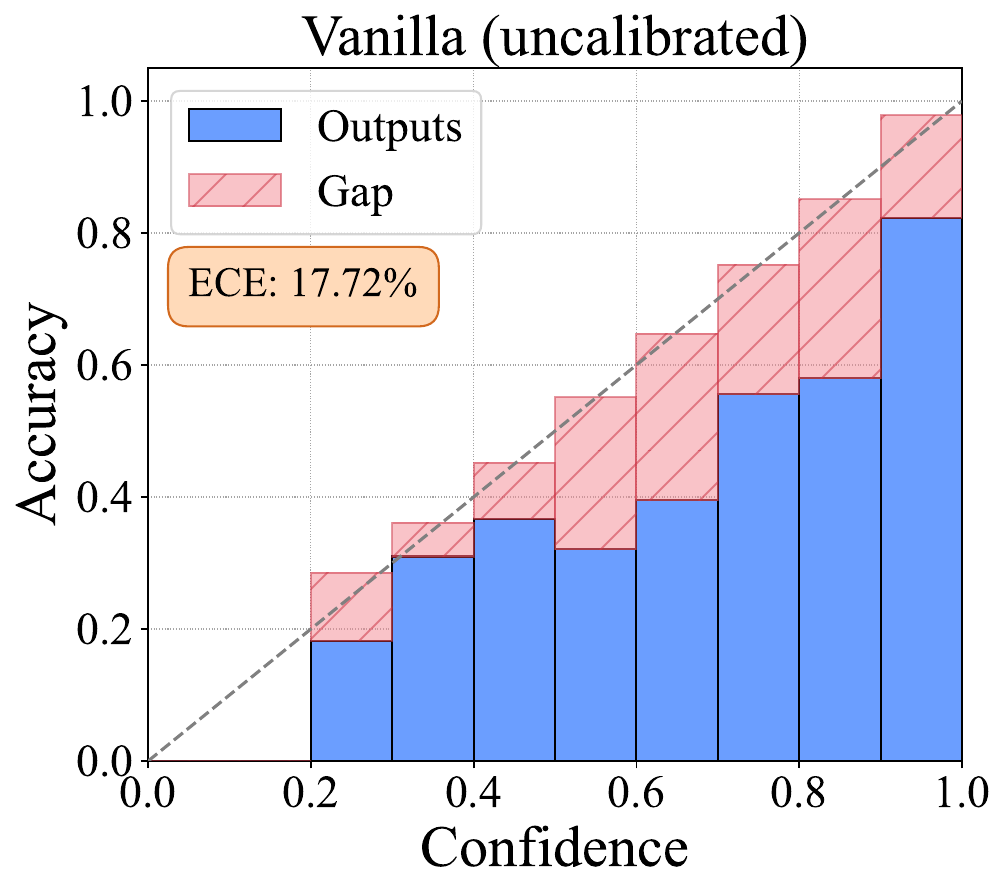}
    \end{minipage}\hfill
    \begin{minipage}{0.24\linewidth}
        \centering
        \includegraphics[width=\linewidth]{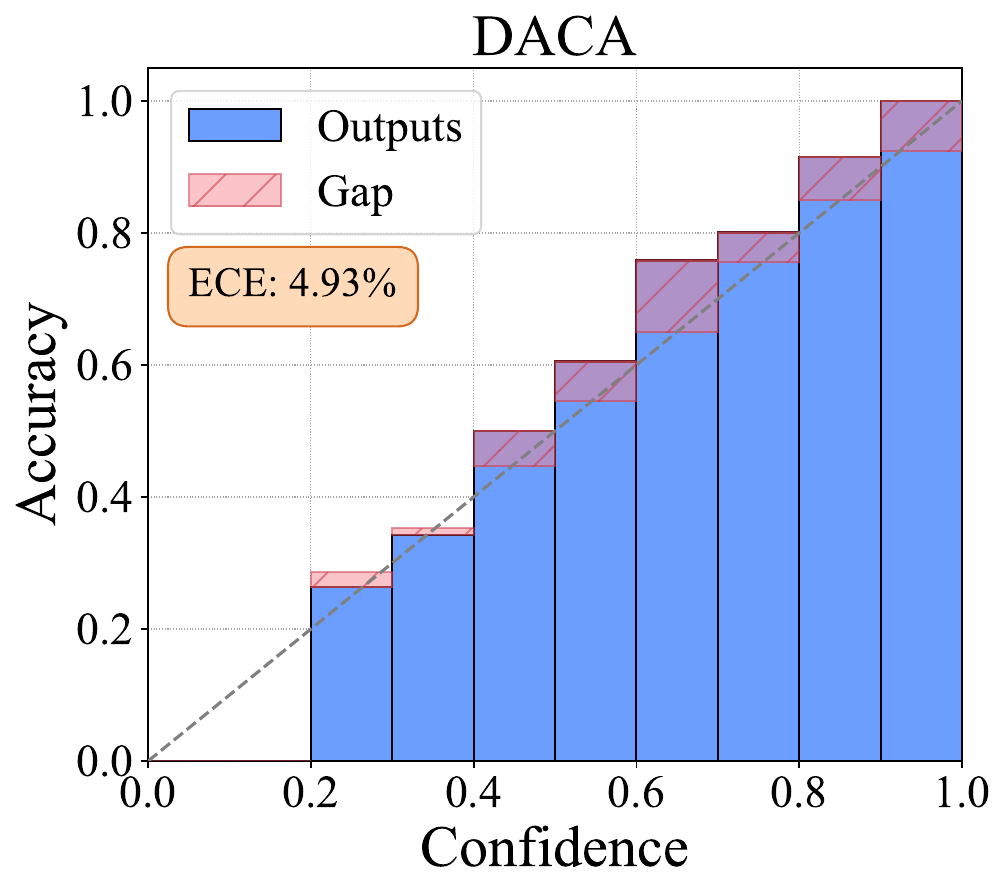}
    \end{minipage}\hfill
    \begin{minipage}{0.24\linewidth}
        \centering
        \includegraphics[width=\linewidth]{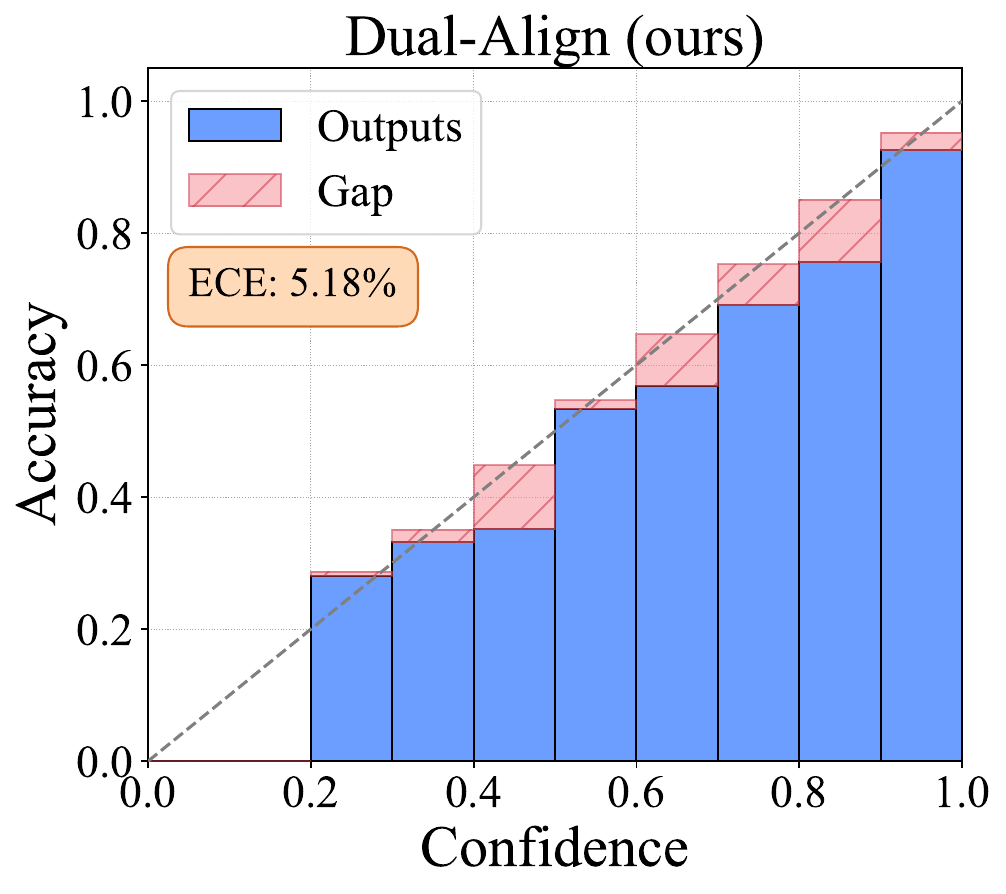}
    \end{minipage}\hfill
    \begin{minipage}{0.24\linewidth}
        \centering
        \includegraphics[width=\linewidth]{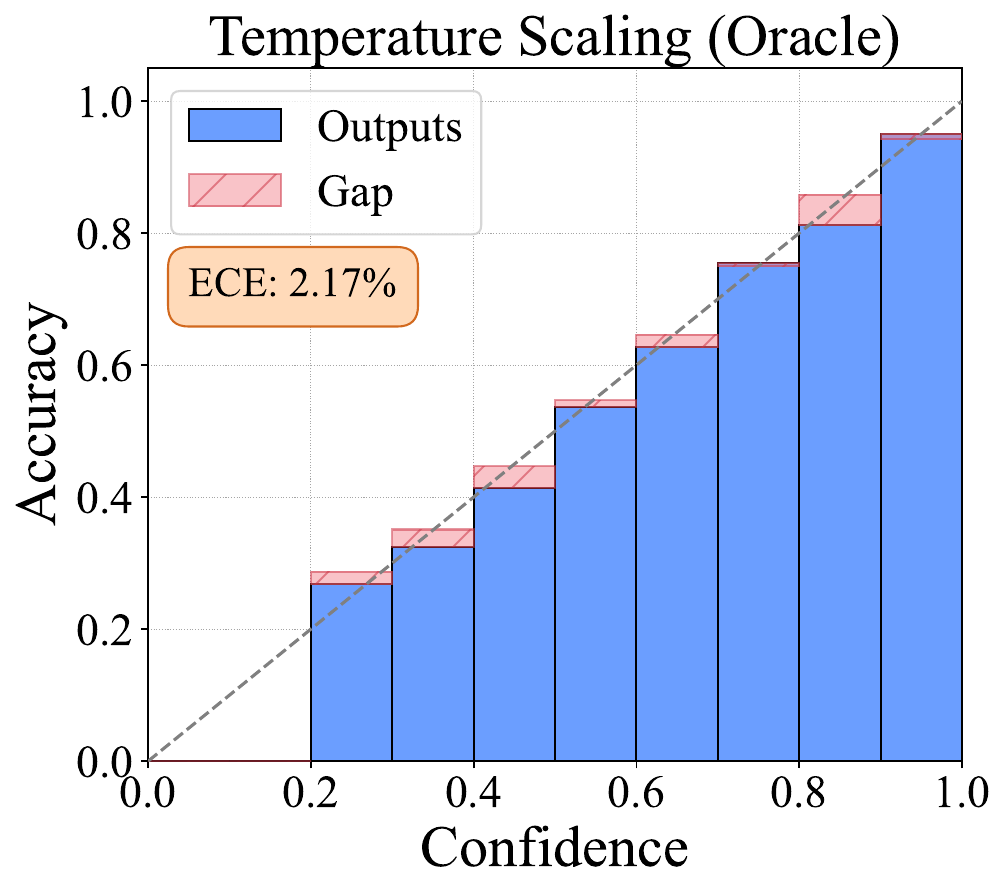}
    \end{minipage}
    \caption{\small Reliability diagrams of Llama3.1-8B-Instruct on MedMCQA dataset.}
    \vspace{-1em}
\end{figure*}

\begin{figure*}[!h]
    \centering
    \begin{minipage}{0.24\linewidth}
        \centering
        \includegraphics[width=\linewidth]{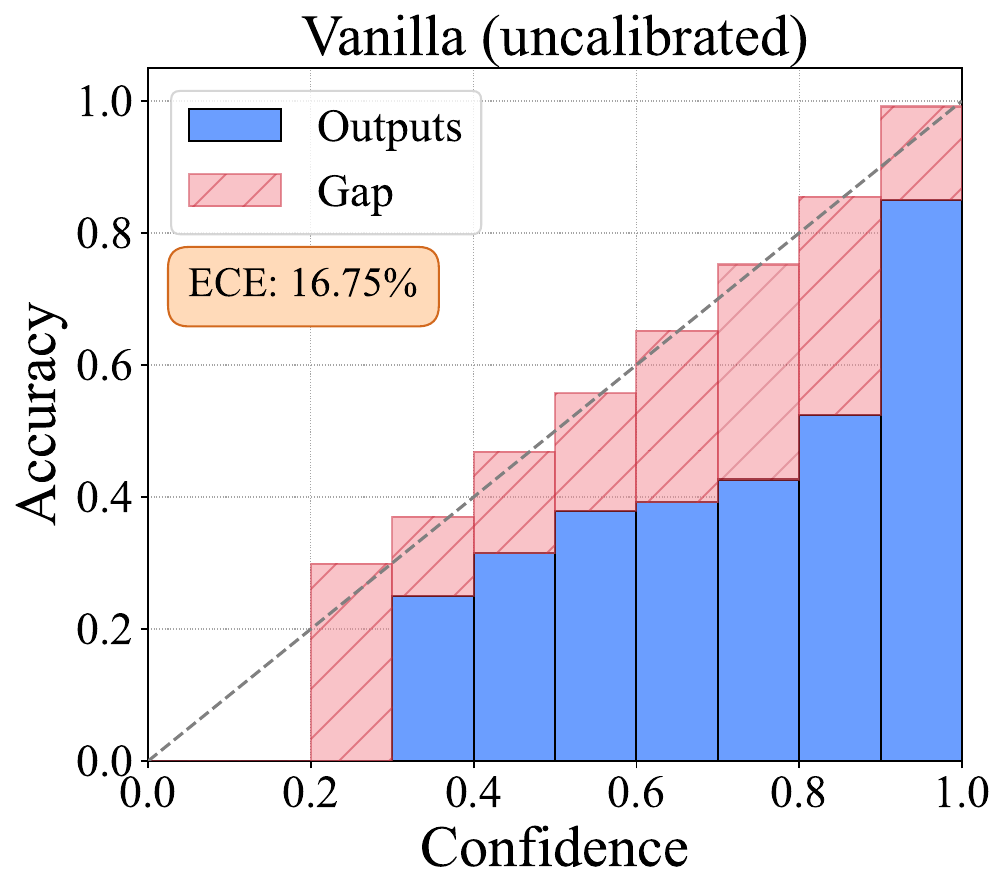}
    \end{minipage}\hfill
    \begin{minipage}{0.24\linewidth}
        \centering
        \includegraphics[width=\linewidth]{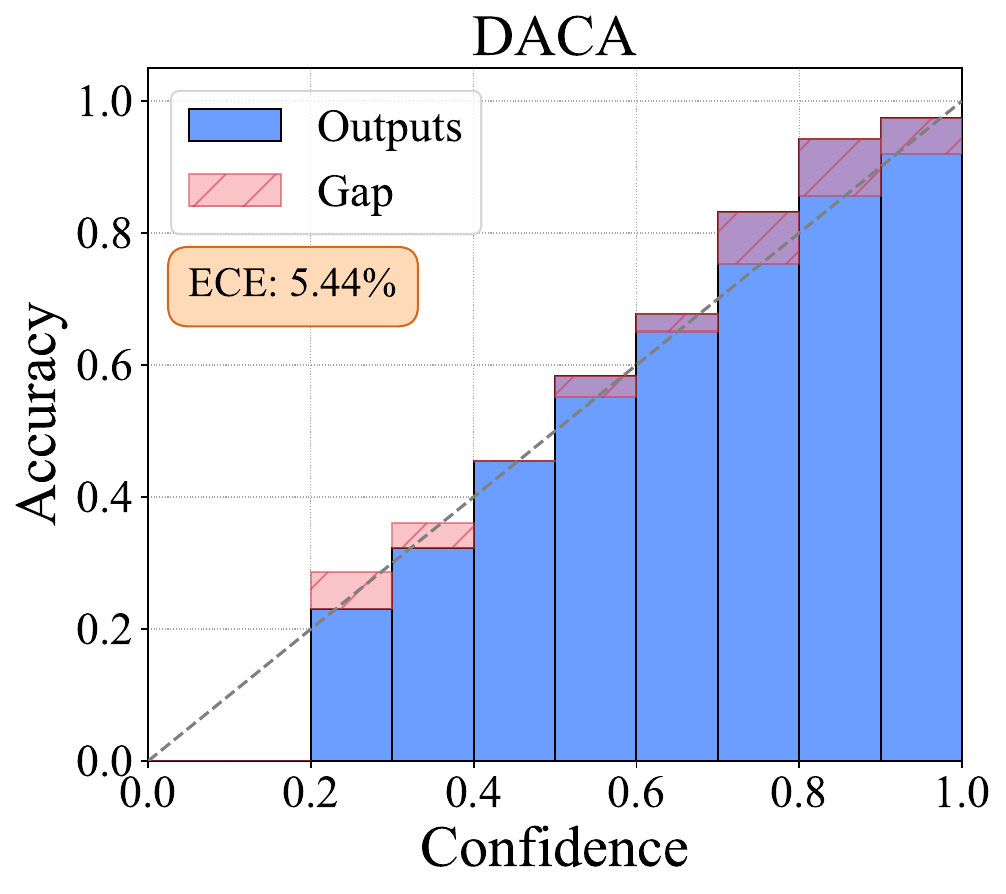}
    \end{minipage}\hfill
    \begin{minipage}{0.24\linewidth}
        \centering
        \includegraphics[width=\linewidth]{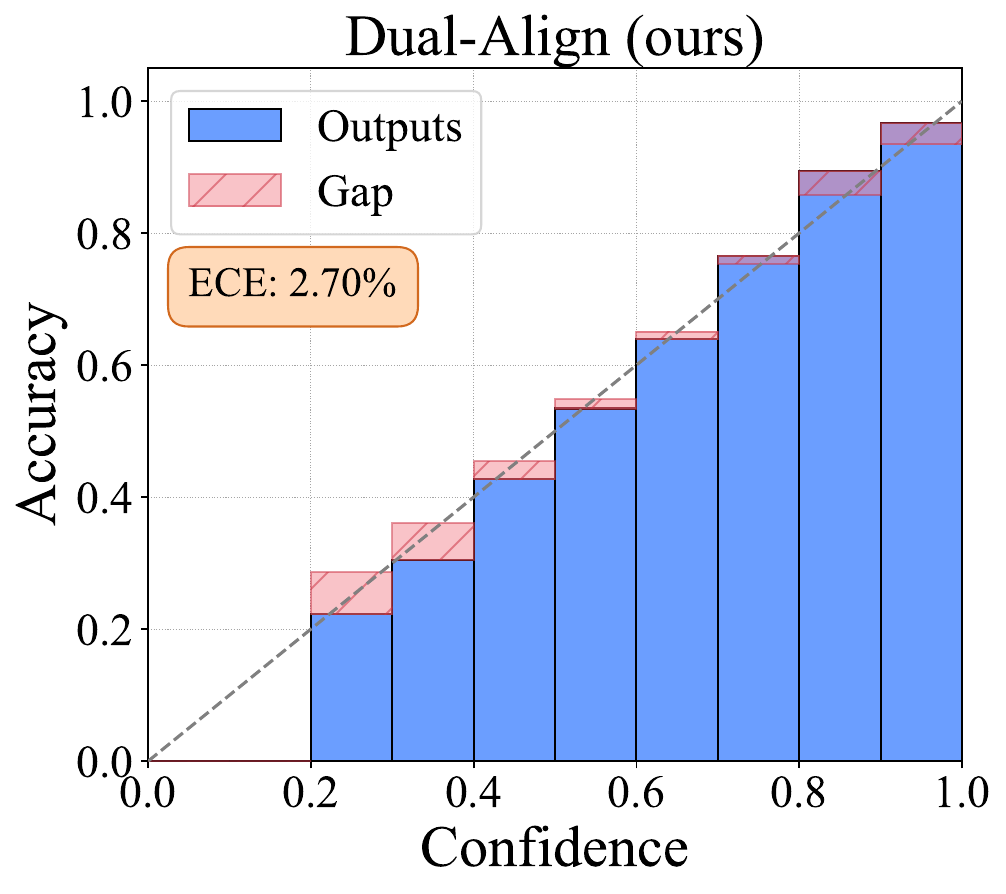}
    \end{minipage}
    \hfill
    \begin{minipage}{0.24\linewidth}
        \centering
        \includegraphics[width=\linewidth]{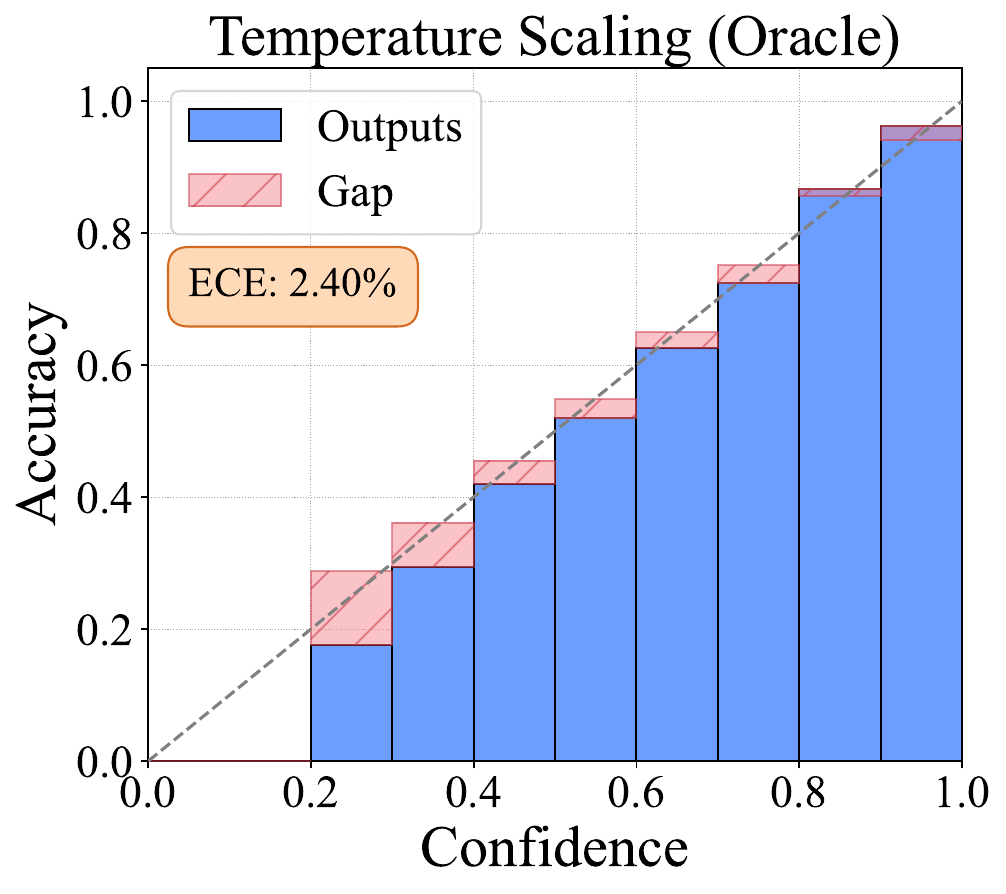}
    \end{minipage}
    \caption{\small Reliability diagrams of Qwen2.5-14B-Instruct on MMLU dataset.}
    \vspace{-1em}
\end{figure*}

\begin{figure*}[!h]
    \centering
    \begin{minipage}{0.24\linewidth}
        \centering
        \includegraphics[width=\linewidth]{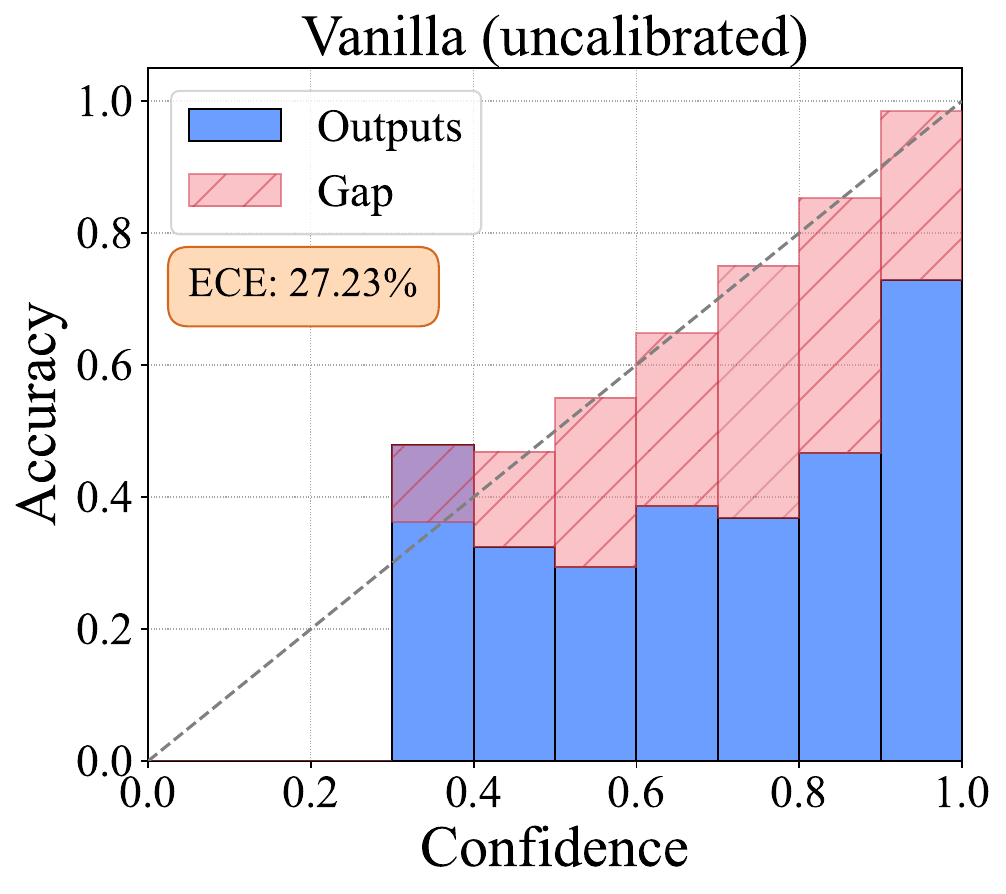}
    \end{minipage}\hfill
    \begin{minipage}{0.24\linewidth}
        \centering
        \includegraphics[width=\linewidth]{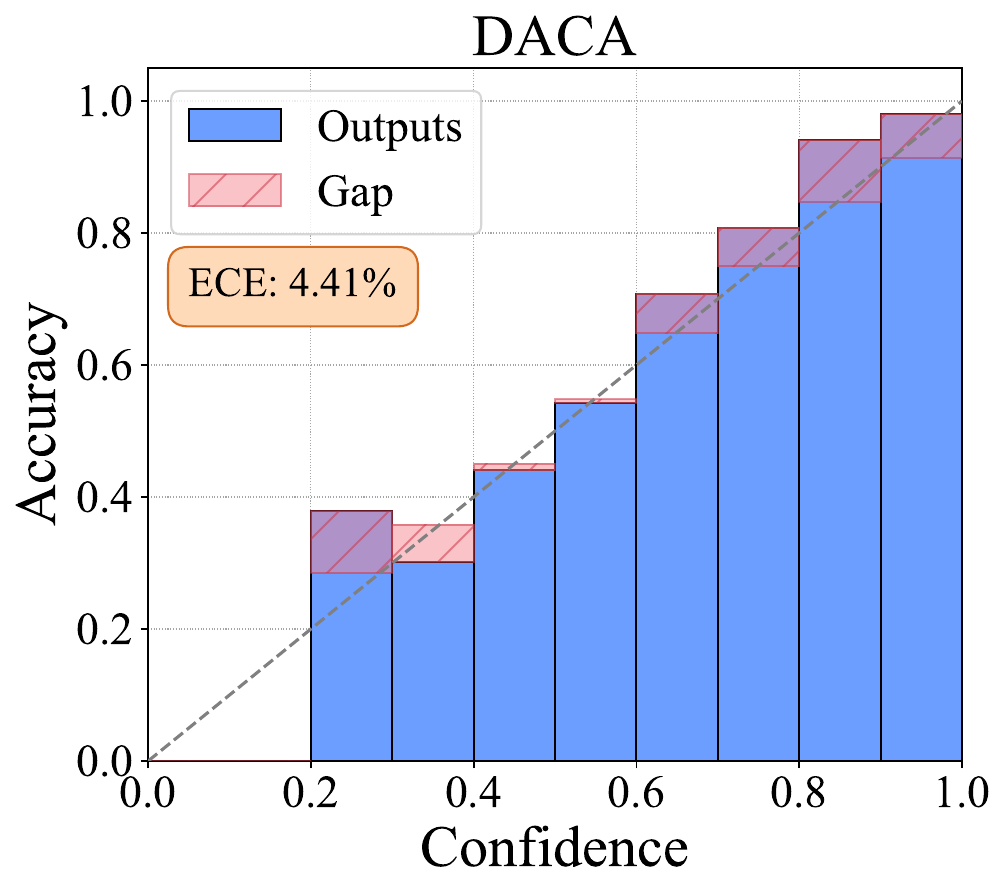}
    \end{minipage}\hfill
    \begin{minipage}{0.24\linewidth}
        \centering
        \includegraphics[width=\linewidth]{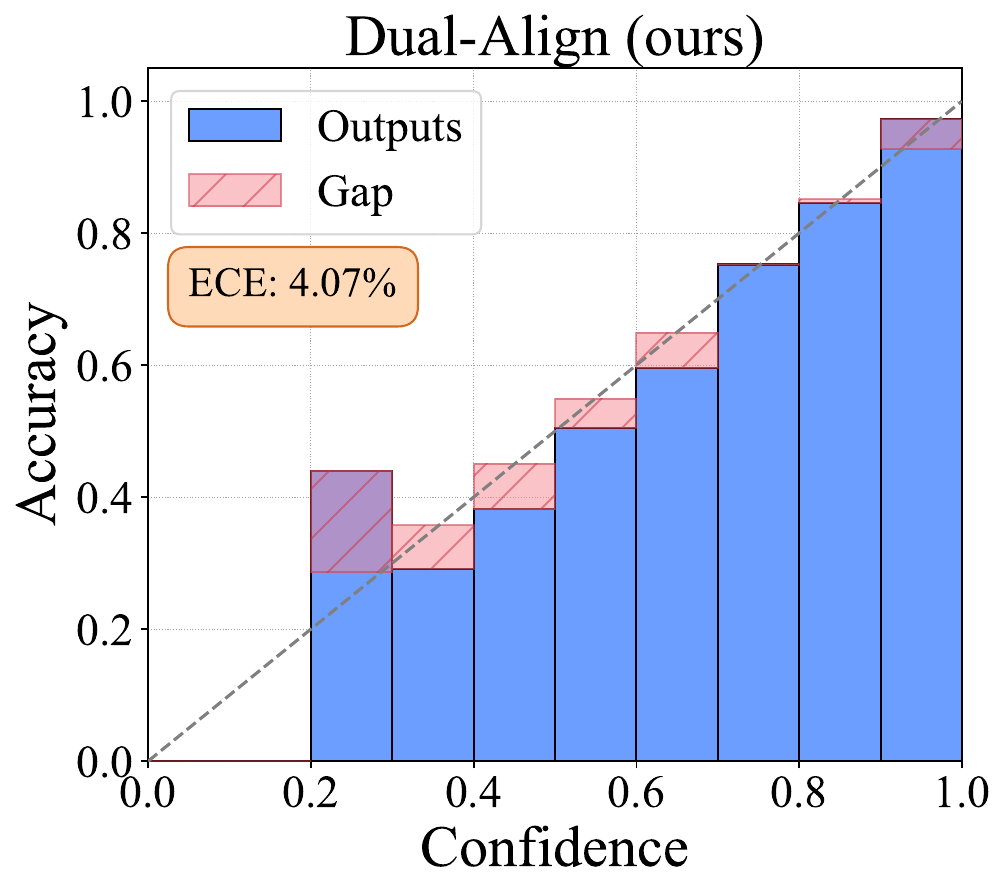}
    \end{minipage}\hfill
    \begin{minipage}{0.24\linewidth}
        \centering
        \includegraphics[width=\linewidth]{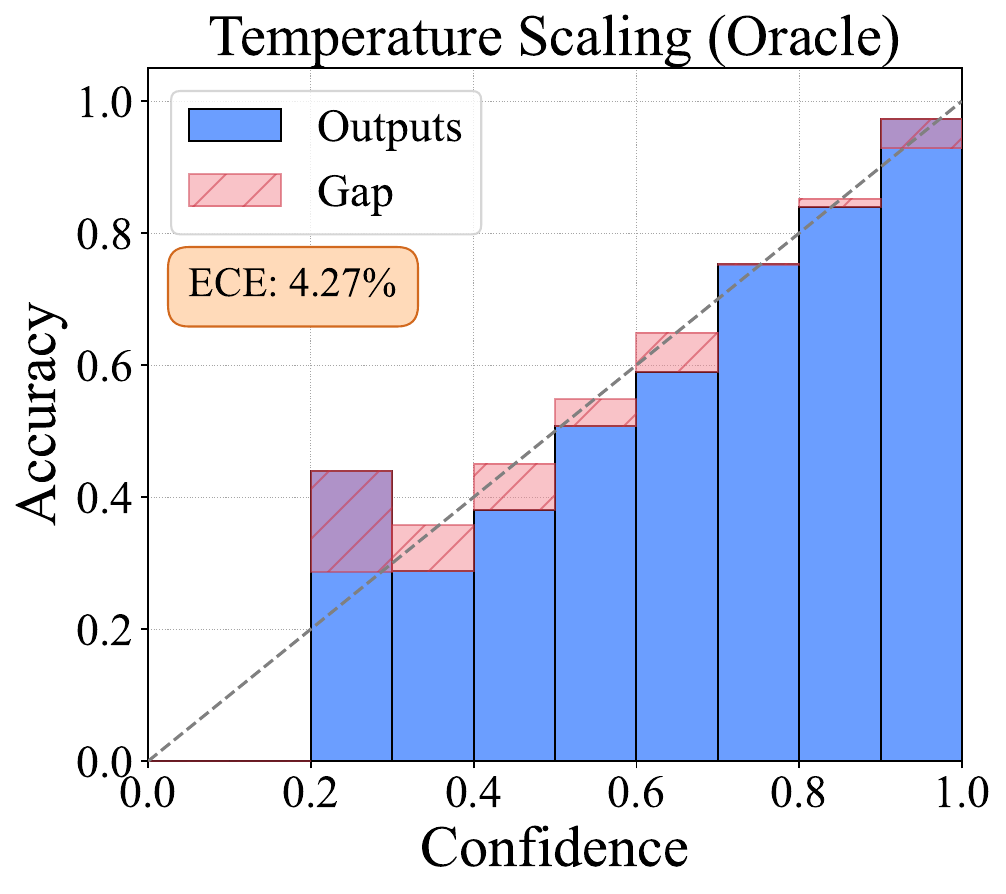}
    \end{minipage}
    \caption{\small Reliability diagrams of Qwen2.5-14B-Instruct on MedMCQA dataset.}
    \vspace{-1em}
\end{figure*}

\begin{figure*}[!htbp]
    \centering
    \begin{minipage}{0.24\linewidth}
        \centering
        \includegraphics[width=\linewidth]{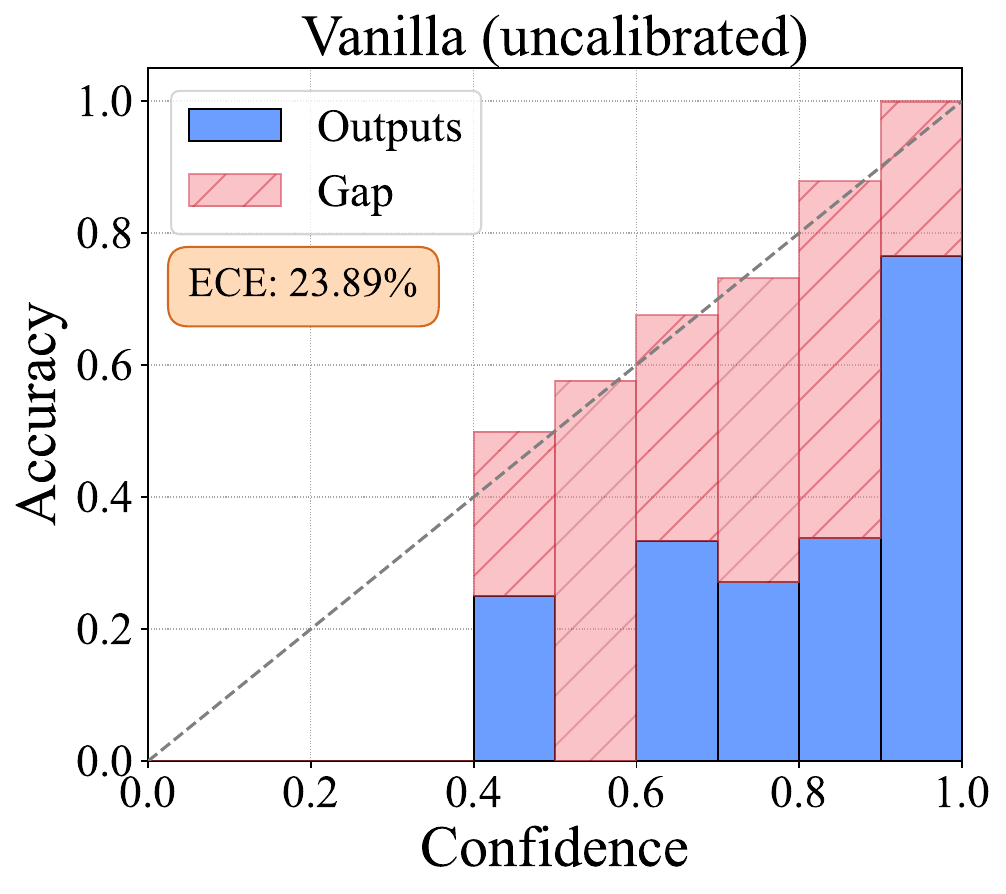}
    \end{minipage}\hfill
    \begin{minipage}{0.24\linewidth}
        \centering
        \includegraphics[width=\linewidth]{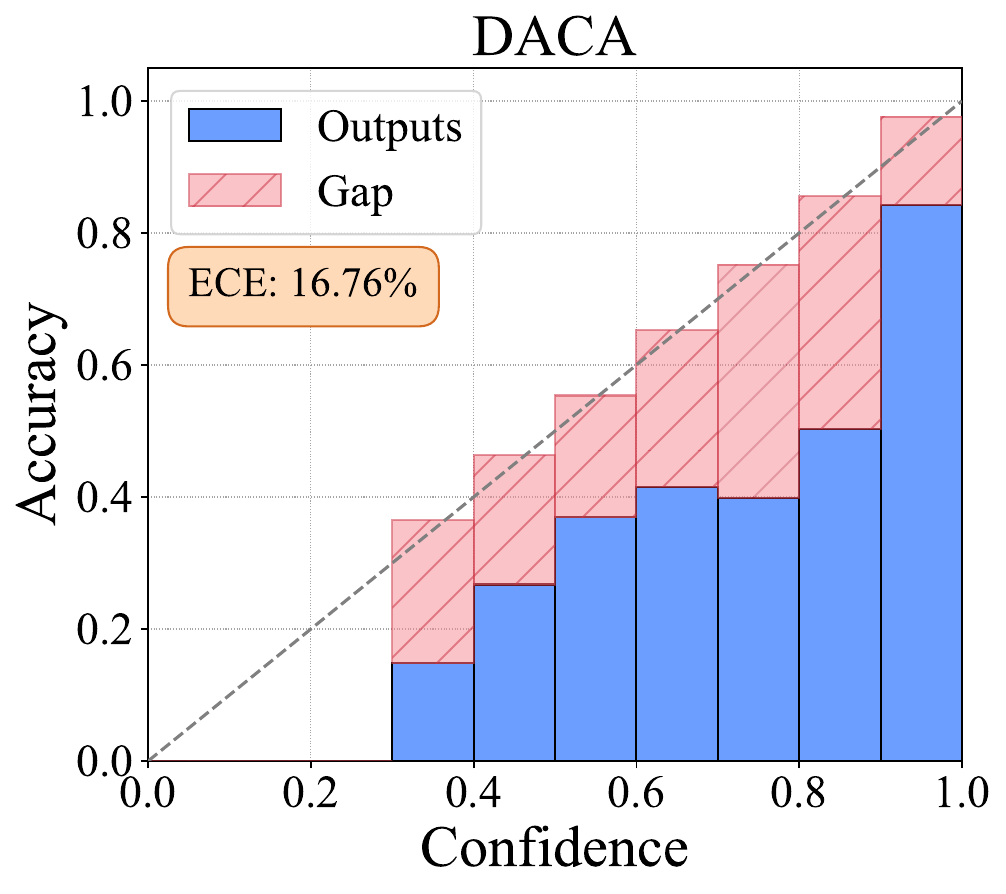}
    \end{minipage}\hfill
    \begin{minipage}{0.24\linewidth}
        \centering
        \includegraphics[width=\linewidth]{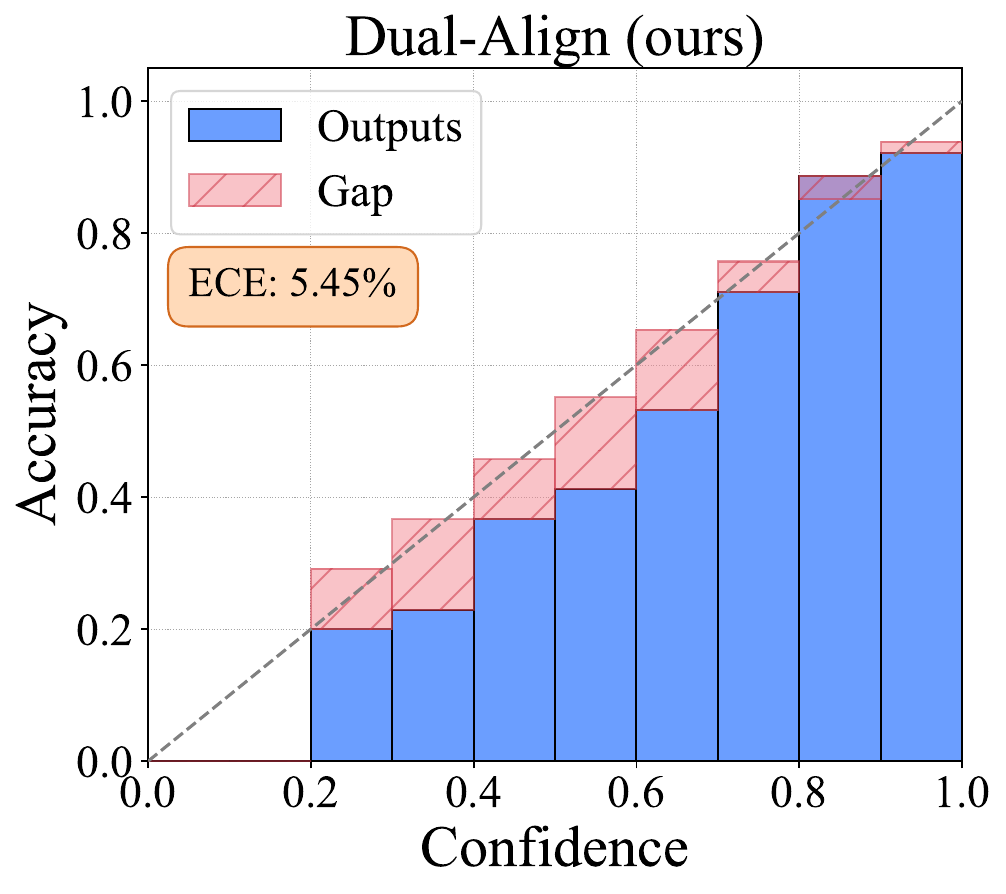}
    \end{minipage}
    \hfill
    \begin{minipage}{0.24\linewidth}
        \centering
        \includegraphics[width=\linewidth]{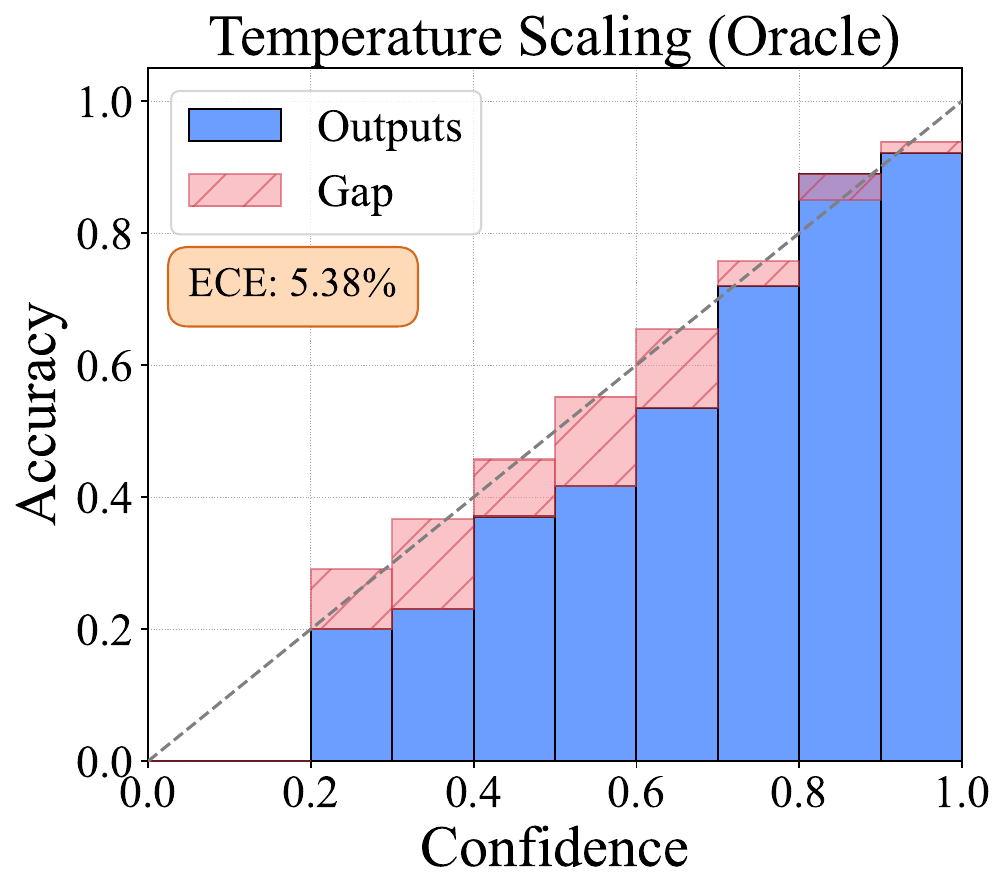}
    \end{minipage}
    \caption{\small Reliability diagrams of Gemma-3-27b-it on MMLU dataset.}
    \vspace{-1em}
\end{figure*}

\begin{figure*}[!htbp]
    \centering
    \begin{minipage}{0.24\linewidth}
        \centering
        \includegraphics[width=\linewidth]{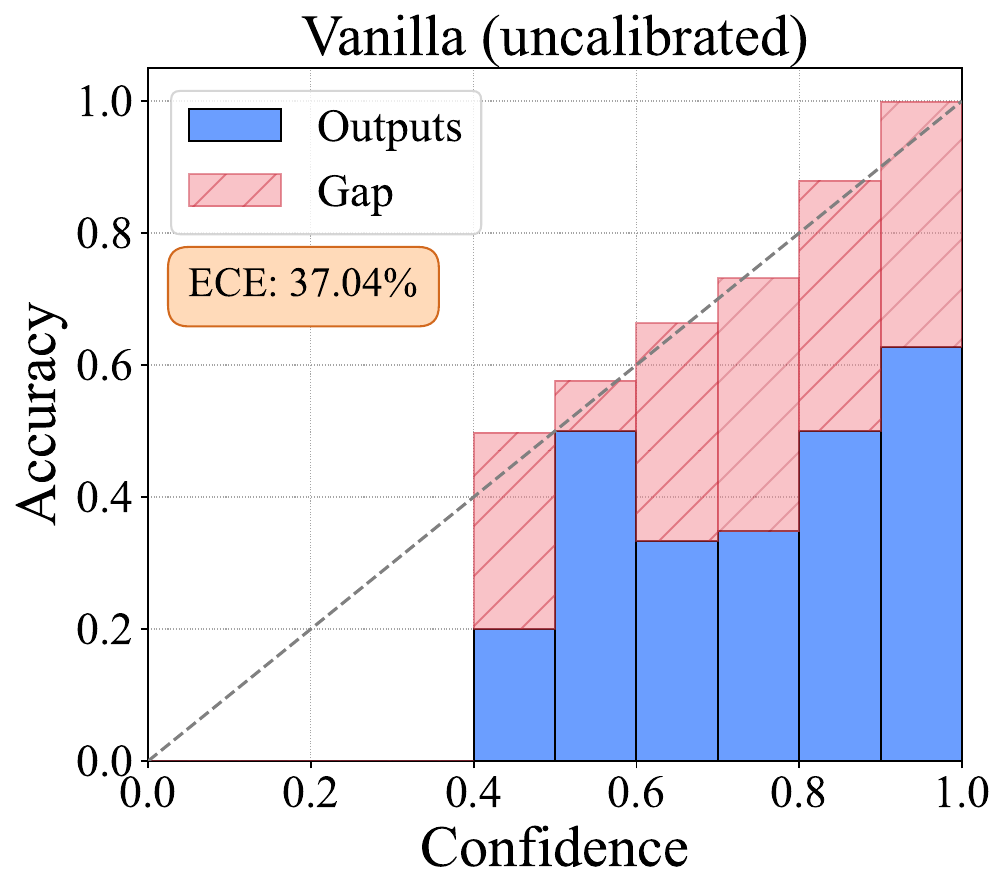}
    \end{minipage}\hfill
    \begin{minipage}{0.24\linewidth}
        \centering
        \includegraphics[width=\linewidth]{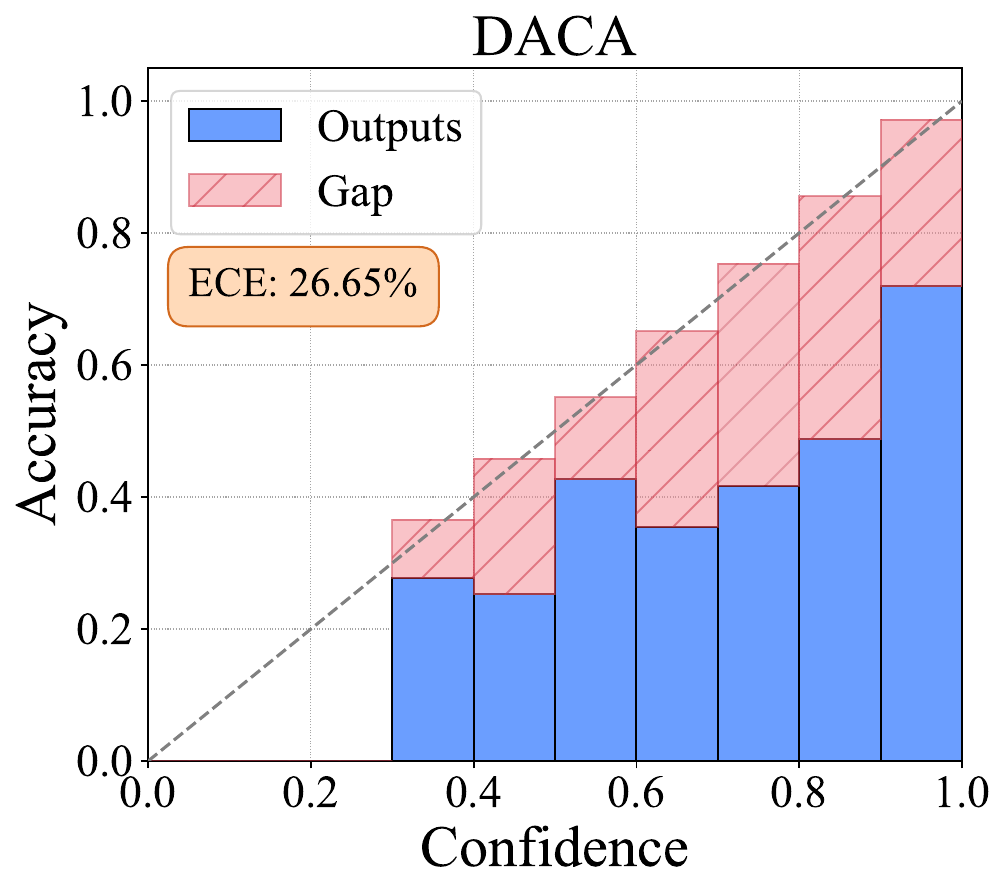}
    \end{minipage}\hfill
    \begin{minipage}{0.24\linewidth}
        \centering
        \includegraphics[width=\linewidth]{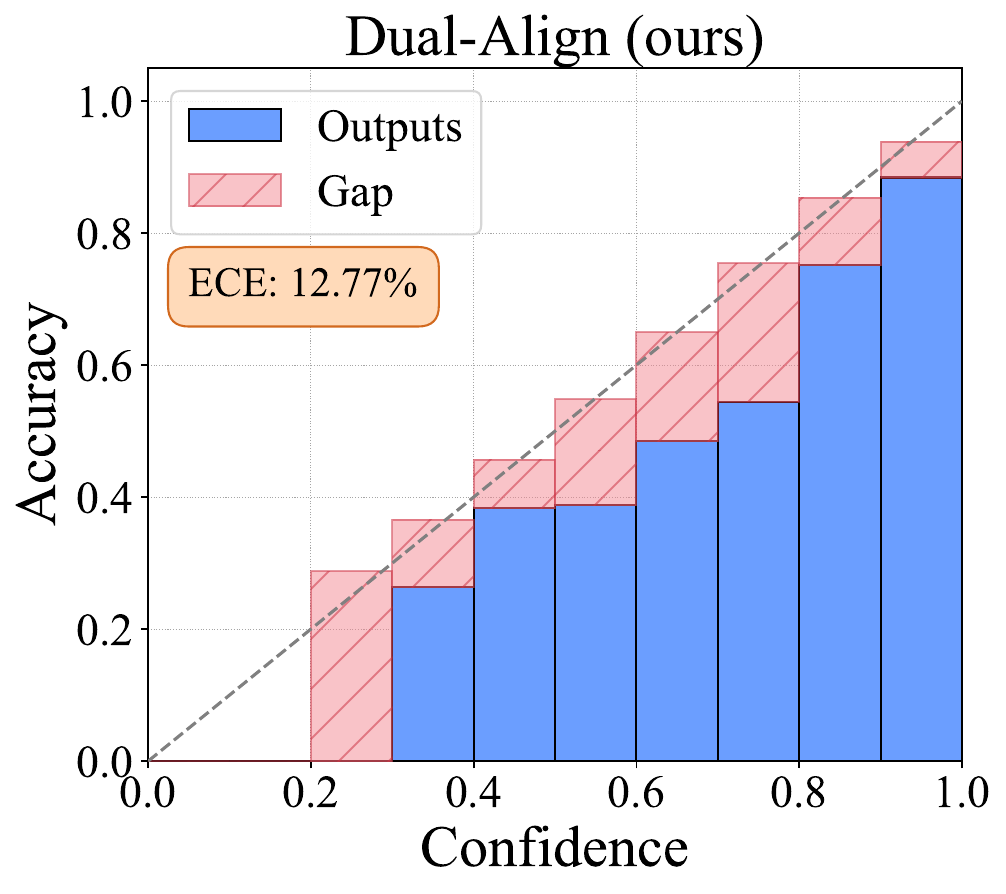}
    \end{minipage}
    \hfill
    \begin{minipage}{0.24\linewidth}
        \centering
        \includegraphics[width=\linewidth]{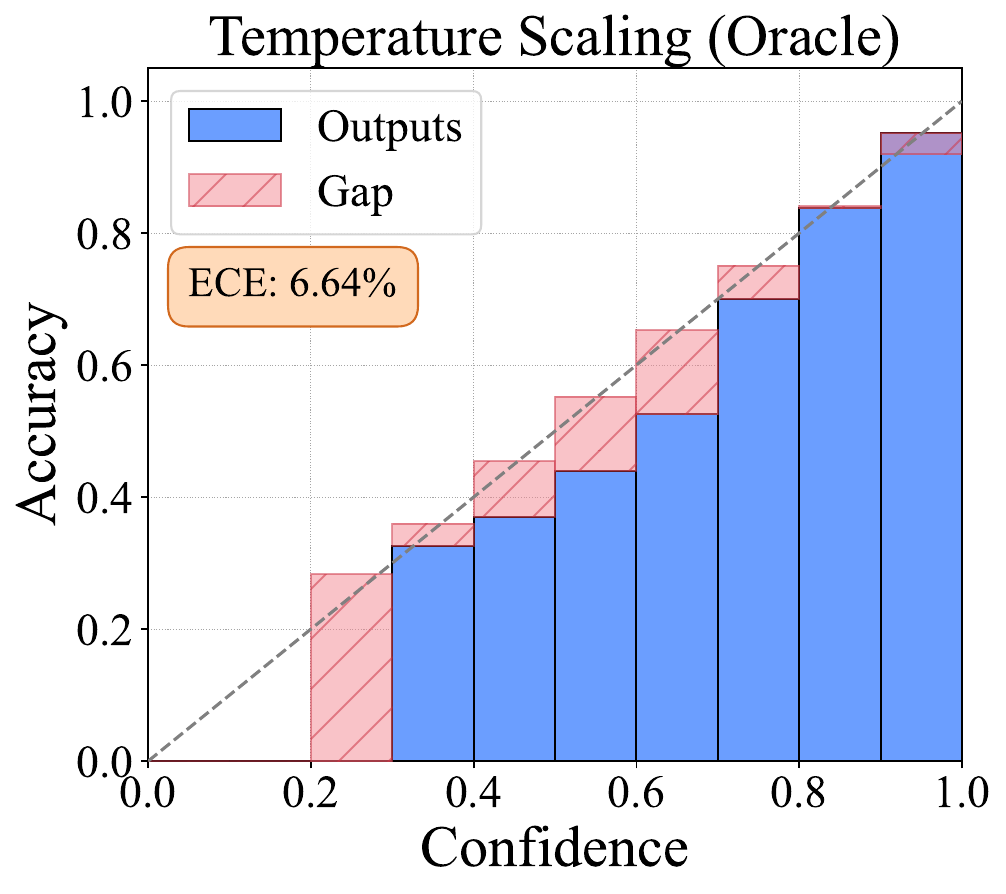}
    \end{minipage}
    \caption{\small Reliability diagrams of Gemma-3-27b-it on MedMCQA dataset.}
    \label{end}
    \vspace{-1em}
\end{figure*}

\end{document}